\documentclass[default,iicol]{sn-jnl}

\usepackage{graphicx}%
\usepackage{multirow}%
\usepackage{amsmath,amssymb,amsfonts}%
\usepackage{amsthm}%
\usepackage{mathrsfs}%
\usepackage[title]{appendix}%
\usepackage[table,xcdraw]{xcolor}
\usepackage{textcomp}%
\usepackage{manyfoot}%
\usepackage{booktabs}%

\usepackage{algorithmicx}%
\usepackage{algpseudocode}%
\usepackage[linesnumbered,ruled,vlined]{algorithm2e}
\usepackage{listings}%
\usepackage{subcaption}
\usepackage{tensor,bm}
\usepackage[inline]{enumitem}
\usepackage{amsmath} 
\usepackage{amssymb}  

\usepackage[inline]{enumitem}
\usepackage{csquotes}
\usepackage{multirow}
\usepackage{booktabs}
\usepackage{subcaption}
\usepackage{tensor,bm}
\usepackage{svg}
\usepackage{algpseudocode}
\usepackage{mathtools}
\usepackage[linesnumbered,ruled,vlined]{algorithm2e}
\definecolor{commentcolor}{rgb}{0.5, 0.5, 0.5} 
\algrenewcommand{\algorithmiccomment}[1]{\hfill \textcolor{commentcolor}{// #1}}

\DeclareMathOperator*{\argmax}{arg\,max}

\theoremstyle{thmstyleone}%
\theoremstyle{thmstyletwo}%

\theoremstyle{thmstylethree}%

\makeatletter
\newcommand{\removelatexerror}{\let\@latex@error\@gobble}
\makeatother
\begin{document}

\title[Article Title]{Navigating the Human Maze: Real-Time Robot Pathfinding with Generative Imitation Learning}


\author*[1]{\fnm{Martin} \sur{Moder}}\email{martin.moder@uni-due.de}

\author[1]{\fnm{Stephen} \sur{Adhisaputra}} 

\author[1]{\fnm{Josef} \sur{Pauli}}

\affil[1]{\orgdiv{Intelligent Systems}, \orgname{University Duisburg-Essen}, \orgaddress{\country{Germany}}}





\abstract{
    This paper addresses navigation in crowded environments by integrating goal-conditioned generative models with Sampling-based Model Predictive Control (SMPC). We introduce goal-conditioned autoregressive models to generate crowd behaviors, capturing intricate interactions among individuals. The model processes potential robot trajectory samples and predicts the reactions of surrounding individuals, enabling proactive robotic navigation in complex scenarios. Extensive experiments show that this algorithm enables real-time navigation, significantly reducing collision rates and path lengths, and outperforming selected baseline methods. The practical effectiveness of this algorithm is validated on an actual robotic platform, demonstrating its capability in dynamic settings.}





\maketitle

\section{Introduction}\label{sec1}

\begin{figure*}[tp]       
    \centering
    \includegraphics[scale=0.47]{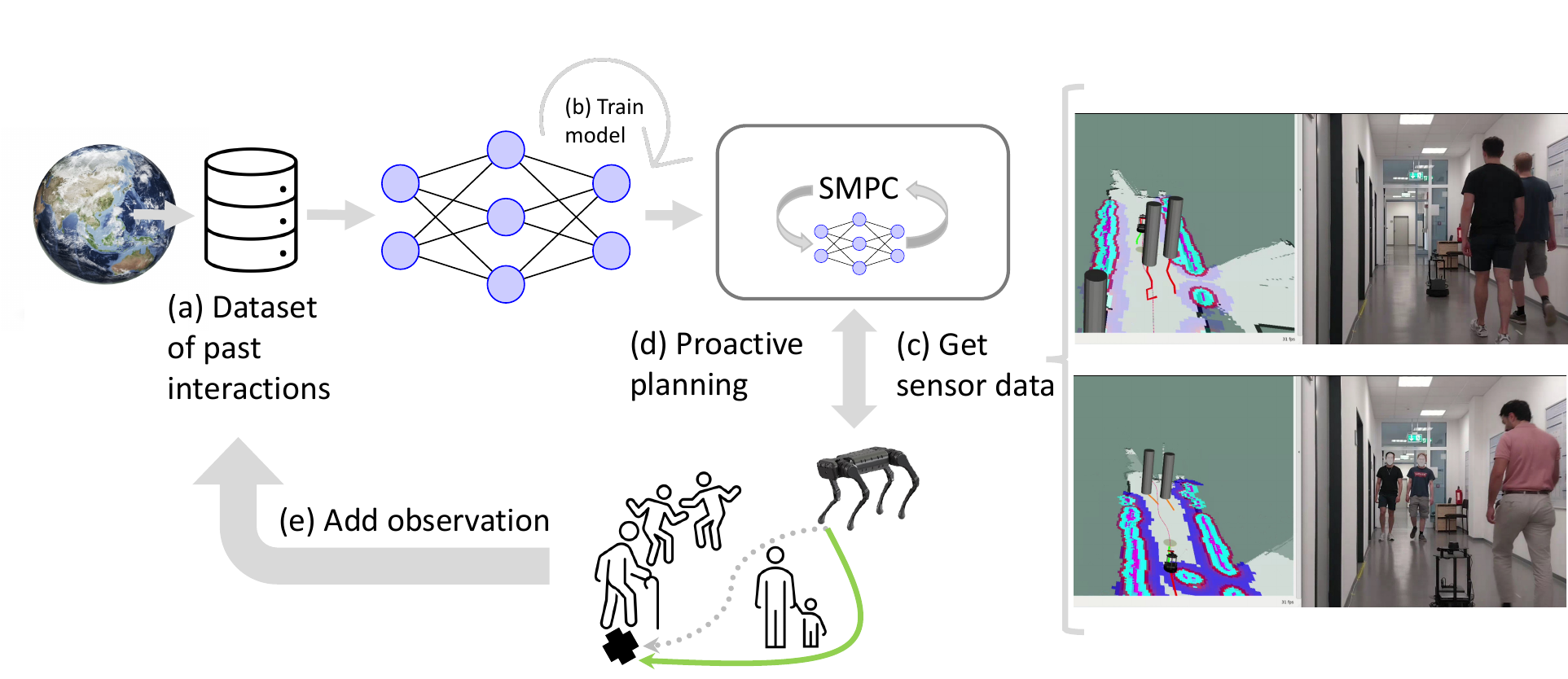}
    \caption{Our approach to imitation learning and planning for robotic navigation in crowded settings is model-based. (a) The dataset comprises recordings of crowd dynamics. (b) Using this dataset, a generative model is trained to forecast future position of individuals. (c) The robot, equipped with a 3D camera and 2D LiDAR sensor, detects and tracks pedestrian positions, and generates a cost map to avoid obstacles. On the left side of the images, four distinct trajectories are shown: agents’ past paths (red), predicted future paths (orange), the robot’s planned trajectory (green), and the robot’s global plan (thin red line). Cylinders represent the positions and outlines of humans. (d) The model predictive control framework, enhanced by the generative model, plans proactively robot trajectories that mimic human movements. (e) New observations can be added to the dataset, allowing the approach to scale with more data. This illustration is adapted from \cite{moder2023MIL}.
    }
    \label{fig:overview}
\end{figure*}

Navigating robots or autonomous cars in crowded areas can lead to “robot freezing,” where the robot becomes stationary due to an unclear path, as noted by Trautman et al. \cite{5654369}. Traditional methods, which only predict crowd behavior without accounting for human-robot interaction, have proven inadequate. Thus, the focus has shifted to strategies promoting cooperative interactions between robots and humans. This has led to the creation of collaborative robots, or “CoBots,” designed to work alongside humans, adapting to both their movements and the complexities of crowded environments.

Reinforcement Learning (RL) is a popular method for CoBots navigation in crowds, offering a comprehensive framework that includes data-driven social acceptance and environment-specific behavior. Despite promising results in simulations and real environments \cite{chen2020relational, everett2021collision}, RL faces challenges due to the need for online crowd interactions, which are costly and safety critical, leading to reliance on unrealistic simulations. Recently, Levine et al. \cite{levine2020offline} have employed offline RL to create task-specific policies using only pre-collected data, eliminating the need for online training. Offline RL requires a reward function to guide behavior, which can be specified during data collection or derived from handcrafted metrics. This is particularly challenging for “social” navigation, where the robot must interpret subtle social behaviors and cues.

Supervised imitation learning, especially using large-scale generative models, shows great promise. These models have made notable contributions in visual recognition \cite{ramesh2022hierarchical} and natural language processing \cite{NEURIPS2020_1457c0d6}. As demonstrated by Cui et al. \cite{cui2022play}, these models can enhance robotic decision-making in complex environments, such as guiding a robotic arm to perform kitchen tasks based on context-driven textual goals.

This work explores the potential of generative models, which are trained on human crowd videos, for cooperative robot action planning. These generative models, we argue, can offer a solution to the "robot freezing" problem by enabling robots to generate intuitive, human-like behaviors, promoting more natural robot-human cooperation. However, applying these models to crowd navigation presents challenges, including processing continuous actions, handling data multi-modality, and conditioning on future outcomes. Additionally, a policy trained solely on human videos won’t match a robot’s unique kinematic and dynamic constraints, and the datasets lack environment representations that robots can easily interpret.

To address these challenges, we propose a hybrid approach combining a likelihood-based generative model, trained on human crowd videos, with SMPC (see Figure \ref{fig:overview}). The generative model, conditioned on goal positions, predicts crowd dynamics as a density function and scores robot plans based on their probability of being human-like. During planning, we use this scoring to create robot plans that mimic human behavior, considering the robot’s physical limits and the natural uncertainty in human decision-making. This hybrid approach enables robots to imitate human behavior in real-world scenarios.

This work builds on \cite{moder2022proactive, collos, moder2023MIL} and extends them in the following ways:
\begin{enumerate}
    \item We advance recent progress in optimal sampling-based planning, focusing on the Model Predictive Path Integral (MPPI) algorithm. Sample-based planning's main advantage is its ability to generate human-like responses to robotic plan samples using a trained generative model. Our research concentrates on the MPPI technique, differing from our previous work \cite{moder2023MIL}, which used the CEM approach. Unlike CEM, which averages the best-sampled trajectories unweighted, MPPI uses a weighted average, enhancing sample selection control and potentially leading to better planning outcomes.
    \item In previous research \cite{moder2023MIL}, we used a generative model to create human-like responses and guide a policy to mimic human behavior closely. However, this model is not goal-conditioned, allowing the robot to exploit predicted human responses, even if it caused humans large detours. By incorporating goal conditioning and a Social Influence Reward (SIR), we limit the robot's ability to exploit human reactions, as individuals follow their own objectives, reducing adaptability to avoid collisions. A goal-conditioned generative model also more effectively directs the robot towards its goal, offering insights on both the human-like nature and efficiency of navigating towards the goal in a crowd.
    \item Current human crowd video data lack an environment representation interpretable by robots, complicating model training to respect static environments. To address this, we use SMPC in robot planning, considering static information during optimization. For generating goal-directed human behavior that respects the environment, we propose a simple yet effective optimization technique to select a sample from the generative model that aligns with the captured environment.
\end{enumerate}

\section{Related Work}

Two pioneering studies on navigation in human environments, RHINO \cite{burgard1999experiences} and MINERVA \cite{thrun2000probabilistic}, both use the Dynamic Window Approach (DWA) \cite{fox1997dynamic} for local collision avoidance, a method still popular in current ROS packages. A key challenge is addressing the unpredictability of human behavior. Du Toit et al. \cite{du2011robot} found that treating agents as independent entities creates overwhelming uncertainty, complicating navigation. Trautman et al. \cite{trautman2015robot} showed that merely constraining this uncertainty, as Du Toit et al. proposed, can cause “robot freezing.” They advocated for a more cooperative approach between humans and robots.

In recent years, CoBots have advanced considerably. Start-ups like Robust AI, Diligent, and Veo Robotics, along with established manufacturers like Kuka and Fanuc, are developing CoBots for harmonious coexistence with humans in shared spaces, with applications ranging from medical navigation to manufacturing assistance. Despite ongoing research, the detailed methodologies of these companies are somewhat elusive. Diligent \cite{Diligent}, for instance, uses “human-guided learning” in hospital operations, where robots learn from their environments and human interactions. However, the integration of imitation learning in their framework is unclear, and their navigation system appears to rely on a classical map-based approach, using QR/APRIL tags at key locations like elevators and doors.

The advancement of CoBots is closely tied to progress in autonomous vehicles. Waymo uses a sense-predict-plan-act pipeline, combining sensor data with maps for environment perception \cite{Gao_2020_CVPR}. Tesla, while following a similar pipeline with a focus on vision, treats lane detection as a linguistic task and employs a unique planning mechanism via tree search \cite{Tesla}. Recently, Tesla has announced a shift towards end-to-end learning. In contrast, Wayve is prioritizing an end-to-end learning pipeline, emphasizing real-time visual inputs over detailed maps \cite{hu2022model}.

\subsection{Learning a Policy}
In most learning-based methodologies, planning occurs in a 2D space, analyzing human dynamics over time and representing social interactions as a comprehensive graph \cite{moder2023MIL}. The focus within this 2D setup is on constructing interaction graphs using neural network structures that handle diverse groups of humans and track their movements over time. Some methods that predict future behaviors provide essential knowledge in this domain \cite{chen2019crowdnav, collos}. Predominant learning approaches include RL and imitation learning, which can operate end-to-end or in a MPC setting. For instance, Chen et al. \cite{chen2017decentralized} pioneered the use of RL to learn a discrete value function suitable for a real robot. Everett et al. \cite{everett2021collision}, using the Actor-Critic paradigm, demonstrated the feasibility of learning a policy for continuous actions. An alternative model-based RL approach is showcased by Chen et al. \cite{chen2020relational}, where their relational graph learns the crowd dynamics model for subsequent tree search.

The referenced RL methods primarily utilize simulations, where human behaviors are often represented using ORCA \cite{ORCA} or the Social Force Model \cite{helbing1995social}. Simulating authentic human behavior in a crowd setting is challenging \cite{CoreCh}, and it becomes even more complex in real-world environments with additional static or dynamic obstacles. To address this issue, some studies focus on imitating expert human behavior using real-world data. For instance, Moder et al. \cite{moder2022proactive,moder2023MIL} employ a likelihood-based generative model to deduce a policy aimed at achieving a set objective while proactively mitigating human interference. Works by \cite{trautman2015robot, trautman2020real} utilize a Gaussian Process model for human interaction prediction, subsequently formulating a robot navigation policy. Most learning-based approaches neglect robot constraints. Of the approaches presented here, only Everett et al. \cite{everett2021collision} account for this by setting dynamics constraints during training, and Moder et al. \cite{moder2023MIL} during optimization in the model predictive control setting.

\subsection{Planning with Generative Models}
Outside the context of crowd navigation, many works have proposed model-based approaches. Common model-based algorithms learn the dynamics model of the world and use it for planning at test time, often through model predictive control and various trajectory optimization methods \cite{chua2018deep, nagabandi2018neural, Rhinehart_2019_ICCV}. The cross-entropy method serves as a practical, sample-based alternative to gradient-based optimization methods, leveraging data-driven dynamics models \cite{chua2018deep, pinneri2021sample, wang2019exploring, moder2023MIL}. Some model-based approaches incorporate a learned policy alongside the dynamics model \cite{wang2019exploring, wu2023daydreamer}, or employ the model to generate “synthetic” samples, enriching the sample set for model-free learning methods \cite{pinneri2020extracting, yu2021combo}.

Another approach, inspired by recent advancements in generative artificial intelligence enabled by transformers \cite{vaswani2017attention}—especially in imitation learning and offline reinforcement learning—is gaining traction. Notably, works that harness transformers in novel ways, diverging from the traditional reinforcement learning paradigm, stand out. For instance, Decision Transformers \cite{decformer} and related methodologies \cite{NEURIPS2021_099fe6b0, srivastava2019training} focus on return-conditioned imitation learning.

\section{Robot Navigation as a Multiplayer Game}\label{sec:problem_f}

Our focus is on a navigation algorithm for an autonomous robot sharing an environment with humans, ensuring the robot is mindful of its impact on human actions. We consider scenarios with $K$ agents: the robot $r \coloneqq 1$ and humans $h \coloneqq \{2,\dots,K\}$. We introduce the necessary variables and derive the general objective.

We define continuous states and discrete time, with agent $k \in K$'s states $\mathrm{S}^{k}_t \in \mathbb{R}^{\Omega_s}$ at time $t$ as 2-dimensional positions on a ground plane ($\Omega_s \coloneqq 2$), with the current time step $t = 0$. The future scene of $K$ agents over $T$ time steps is $\mathrm{S}^{1:K}_{1:T} \in \mathbb{R}^{T \times K \times \Omega_s}$. Let $\mathrm{S}^r_t \coloneqq \mathrm{S}^1_t \in \mathbb{R}^{\Omega_s}$ be the state of the robot, and $\mathrm{S}^h_t \coloneqq \mathrm{S}^{2:K}_t \in \mathbb{R}^{(K-1) \times \Omega_s}$ the states of all humans. Absence of a time step subscript denotes all future or past time steps, and absence of an agent index superscript denotes all agents, e.g., $\mathrm{S} \coloneqq \mathrm{S}^{1:K}_{1:T}$. Capital roman letters denote random variables, with realizations in roman lowercase. For instance, past states of all agents over a period \(T_o\) are \(\mathrm{o} := s_{-T_o:0}\).

The next future states \(\mathrm{S}_{t+1}\) of all agents, determined by their actions \(\mathrm{A}_t \coloneqq \mathrm{A}^{1:K}_t \in \mathbb{R}^{K \times \Omega_s}\), use two transition functions: \(f_r\) for robot dynamics and \(f_h\) for human dynamics. Continuous actions \(\mathrm{A}_t^r\) at time \(t\) are decided by a stochastic robot policy:
\begin{equation}
    \mathrm{A}^r_t \sim \pi^r(\cdot |\mathrm{S}_t;\theta_{\pi^r}),
\end{equation}
where $\theta_{\pi}$ represents the distribution parameters.

The robot's state is influenced by its actions as defined by the robot dynamics function:
\begin{equation}\label{eq:fr}
    \mathrm{S}^r_{t+1} = f_r(\mathrm{S}^r_{t}, \mathrm{A}^r_t).
\end{equation}
Human actions \(\mathrm{A}^h_t \coloneqq \mathrm{A}^{2:K}_t\) are decided by a human policy \( \pi^h:\mathbb{R}^{(K-1) \times {\Omega}} \to \mathbb{R}^{(K-1) \times \Omega_s}\), detailed in the next section. The human transition function is:
\begin{equation}\label{eq:fh}
   \mathrm{S}^h_{t+1} = f_h(\mathrm{S}^h_{t}, \mathrm{A}^h_t) = \mathrm{S}^h_{t} + \mathrm{A}^h_t.
\end{equation}

The objective is to determine the optimal parameters, \( \theta_{\pi^r}^* \), for the robot's policy \( \pi^r \), to maximize a specified scalar return \( R \in \mathbb{R} \). Considering \( f_R(\tau):\mathbb{R}^{T \times K \times (\Omega_s + \Omega_s)} \to \mathbb{R} \) as the finite-horizon undiscounted return function, the expected return is:
\begin{equation}\label{eq:reward}
    \begin{aligned}
        J(\pi^r, \pi^h) &= \mathbb{E}_{\tau} \left[f_R(\tau)\right] \\
        &= \mathbb{E}_{\tau} \left[ \sum_{t=0}^{T-1} f_\phi(\mathrm{S}_t, \mathrm{A}^r_t, \mathrm{A}^h_t) \right]
    \end{aligned}
\end{equation}
where \( \mathbb{E}_{\tau}\left[\cdot\right] \) denotes the expectation over the episode \( \tau = \{ s_0, \mathrm{A}^r_0, \mathrm{A}^h_0, \mathrm{S}_1, \cdots, \mathrm{S}_{T-1}\} \), starting from the current observed states \( s_0 \) for all agents. The function \( f_\phi(\mathrm{S}_t, \mathrm{A}^r_t, \mathrm{A}^h_t):\mathbb{R}^{K \times (\Omega_s + \Omega_s)} \to \mathbb{R} \) represents the robot's reward function, detailed in Section \ref{sec:reward}. The robot aims to find the optimal parameters to maximize the expected return:
\begin{equation}\label{eq:mpc_obj}
    \theta_{\pi^r}^* = \argmax_{\theta_{\pi^r}} J(\pi^r, \pi^h).
\end{equation}
To solve the finite horizon problem as stated in \eqref{eq:mpc_obj}, the robot requires knowledge of $\pi^h$, implying the necessity to understand human cognition and predict human responses under various scenarios. 

This challenge is often circumvented by assuming the human policy is based solely on human states, ignoring the robot's state. This assumes humans continue their trajectory at their current velocity as if the robot were invisible. Consequently, the robot acts as if it has no influence on the environment, planning movements within the confines of existing free space. However, this can lead to 'robot freezing,' where the robot remains stationary, waiting for natural changes in the environment to present more free space.

An alternative strategy approximates the human reward function using Inverse Reinforcement Learning \cite{sadigh2016planning}. This derives reward functions for both the robot and humans, facilitating the formulation of individual policies, such as through MPC. Each agent, human or robot, aims to maximize their own returns subject to dynamic constraints. However, the interdependence of reward functions—where each agent's reward depends on the states and actions of all other agents—and potential conflicts between individual reward functions, elevate the problem to the domain of game theory. Identifying policies that optimize the return for all agents simultaneously may be infeasible. In subsequent sections, imitation learning via Behavior Cloning (BC) is proposed to make the objective in \eqref{eq:mpc_obj} manageable.

\section{Methods}\label{sec:gcbc}


In \eqref{eq:mpc_obj}, the objective is framed as a multiplayer game, where each agent's actions are influenced by others, leading to complex interactions. Unlike the traditional RL goal of finding a single globally optimal policy, MPC iteratively seeks locally optimal parameters for the robot's policy. This process uses a predictive model to project future environmental states over a finite horizon \( T \), incorporating the robot's policy and transition dynamics, as specified in \eqref{eq:fr}, along with the human policy and its transition dynamics, as detailed in \eqref{eq:fh}.

To address the complexities in \eqref{eq:mpc_obj}, a goal-conditioned human density \( p(\mathrm A^h_t | \mathrm S_{\leq t}, \mathrm G^h, \mathrm o; \theta_{nar}) \) is introduced. This density captures human control actions \( \mathrm A^h_t \) conditioned on all observed states of all agents $\mathrm S_{\leq t} \coloneqq \mathrm S_{1:t}^{1:K}$ and their goals \( \mathrm G^h \in \mathbb{R}^{\Omega_s}\). The computation employs a NAR model, as specified in \eqref{eq:ar_h_policy}. The parameters, \( \theta_{nar} \), are optimized to enhance the likelihood of the observed data, an approach also referred to as Goal Conditioned Behavior Cloning (GCBC). Consequently, the human policy is formalized as:
\begin{align}\label{eq:human_p}
    \pi^{h}_{nar}(\mathrm A^h_t & \mid \mathrm S_{\leq t}, \mathrm G^h, \mathrm o; \theta_{nar}) \notag \\
    &\coloneqq p(\mathrm A^h_t \mid \mathrm S_{\leq t}, \mathrm G^h, \mathrm o; \theta_{nar}).
\end{align}

\begin{figure}[t]
    \centering
    \includegraphics[scale=0.063]{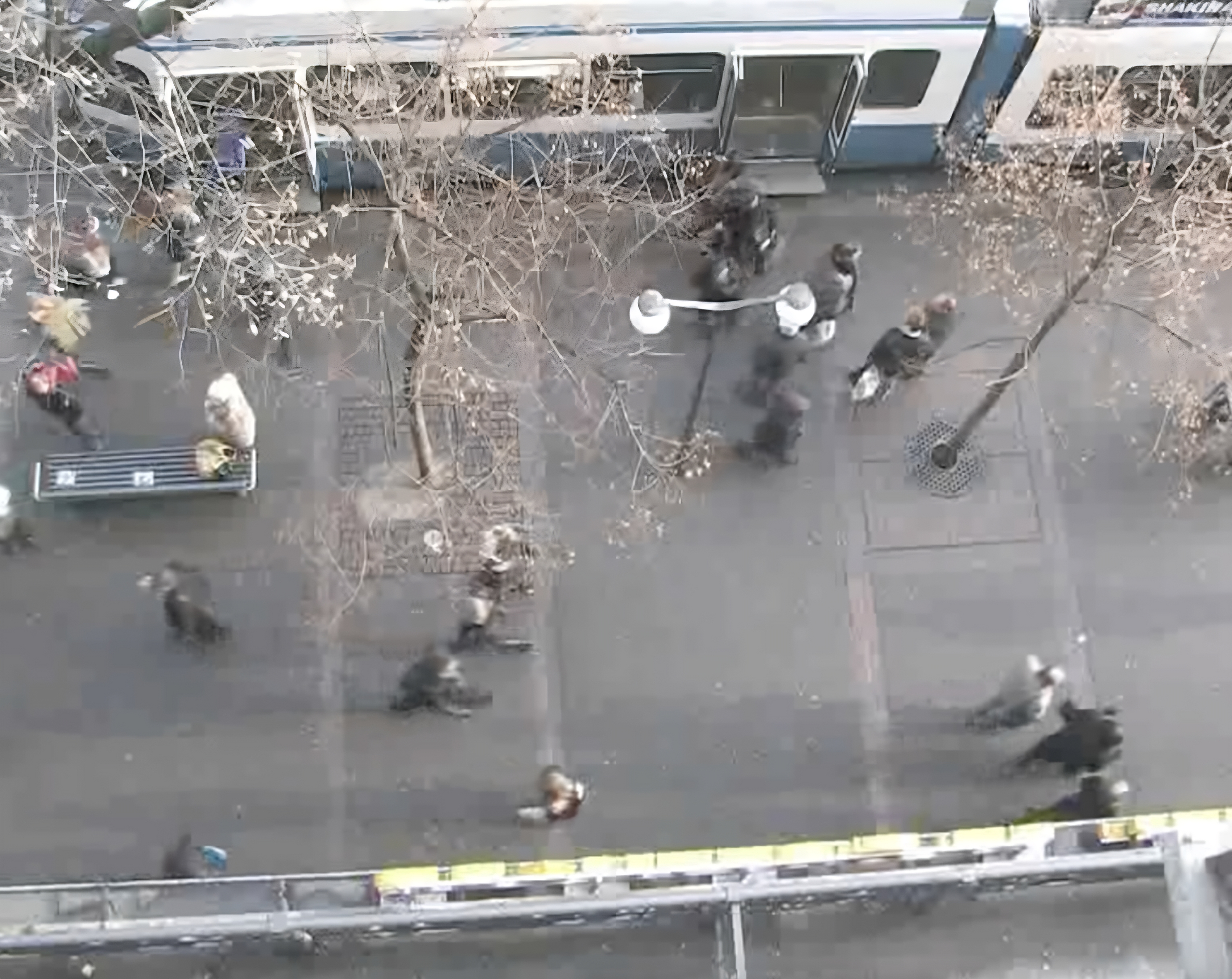}
    \caption{A snapshot from the second sequence in the ETH pedestrian dataset \cite{pellegrini2010improving}, showcasing a crowded street scenario.}
    \label{fig:eth}
\end{figure}

The dataset contains human positions extracted from real-world video recordings, as shown in Figure \ref{fig:eth}. To make this dataset compatible with GCBC, goal relabeling of future states is necessary. A common data augmentation technique, useful when the set of goals $\mathrm G^h$ is a subset of the observation space $\mathrm S^h$, is Hindsight Experience Replay (HER) \cite{andrychowicz2017hindsight}. This approach augments the dataset with additional goal information, as follows:
\begin{equation}
    \begin{split}
        \mathcal{D}_{\text{her}} \coloneqq \bigg\{ \bigg( &\tensor*[^m]{\mathrm{s}}{^{h*}_{T}}, \tensor*[^m]{\mathrm{o}}{^{h*}}, \\
        &\{ ( \tensor*[^m]{\mathrm{a}}{^{h*}_{t-1}}, \tensor*[^m]{\mathrm{s}}{^{h*}_{t}} )\}_{t=1}^{T} 
        \bigg) \bigg\}_{m=1}^{M_{\text{her}}} \,,
    \end{split}
\end{equation}
where $*$ denotes expert actions and states, and $M_{\text{her}}$ is the number of scenes captured.


With the human policy defined, a response to a sequence of robot actions \( \mathrm a^r \coloneqq \mathrm a^r_{0:T-1} \) aimed at achieving a robot goal \( \mathrm g^r \) can be computed. This action sequence is converted into robot states \( s^r \) via the transition function \( f_r \). The human policy \( \pi^{h}_{nar} \) then generates an autoregressive response, considering forecasted human goals \( \mathrm g^h \). At each time step \( t \), the robot state \( \mathrm s_t^r \) and observed human states \( \mathrm s_t^h \) are concatenated to create a joint state \( s_t = s_t^r \oplus s_t^h \). The human policy applied to this joint state derives the human action \( \mathrm a^h_t \), leading to the next human state \( \mathrm s^h_{t+1} \) through \( f_h \). This autoregressive process over the horizon \( T \) yields the human response sequence \( \mathrm a^h \) to the robot's actions. The reward for these human responses is computed as defined in \eqref{eq:reward}. In this framework, the robot engages in a 'multiplayer game,' eliciting human policy responses to its actions. This approach, \textit{Best-response Iteration}, is based on game theory and is noted for its efficacy in recent research \cite{sadigh2016planning, williams2018best}, including this study.

In goal conditioning, each individual is assumed to aim for a specific location. This influences the robot's navigation plan in two ways: by accounting for interaction dynamics in crowded spaces and by aligning with individual human goals. Consequently, the robot effectively navigates towards its own goal while acknowledging human objectives. The robot's plan assumes that humans are proactive to an extent; their goal-driven actions make them somewhat predictable but not overly flexible.

Moder et al. \cite{moder2023MIL} demonstrated that human policies trained with GCBC can support effective robot planning. However, this assumes the robot can emulate human movement, a challenge given the stricter kinematic and dynamic constraints on robots. Additionally, no comprehensive dataset exists that captures both human positions and static obstacles on a large scale, limiting GCBC to human-populated environments without static obstacles. Due to the difficulty of simulating such data, a hybrid strategy is adopted. This strategy uses SMPC and spatial mapping for the robot's local plan while the human policy predicts human actions. The human policy also critically evaluates the robot's proposed plans, identifying those that best emulate human-like behavior.

\subsection{Sampling-Based Model Predictive Control}

The SMPC methodology is based on importance sampling. The robot’s next action is determined by an optimal policy, which is unknown and cannot be directly sampled. However, samples can be evaluated using the reward function $f_\phi$. The objective is to refine the known policy so its samples approximate the optimal distribution. An accurate reward function is essential for assessing the samples, as outlined in \eqref{eq:mpc_obj}. Here, the robot policy for SMPC in the robot action space is described as a Gaussian density:
\begin{equation}\label{eq:mppi_p}
    \pi^r_{gauss}( \mathrm{\acute A^{r}_t}; \mu_{t}, \Sigma_{t}) = \mathcal{N}(\mu_{t}, \Sigma_{t}),
\end{equation}
where \( \mu_{t} \) and \( \Sigma_{t} \) are the mean and covariance at timestep \( t \), respectively, defined in \(\mathbb{R}^{\Omega_s}\) and \(\mathbb{R}^{\Omega_s \times \Omega_s}\). 
In \eqref{eq:mppi_p}, policy actions are represented as linear and angular velocities, denoted by \(\mathrm{\acute{A}^{r} }\coloneqq \mathrm {\acute{A}^{r}_{0:T-1}}\), where \(\mathrm{\acute{A}^{r}_{t}} \in \mathbb{R}^{\Omega_s}\) and \(\mathrm{\acute{A}^{r}_{0:T-1}} \in \mathbb{R}^{T \times \Omega_s}\). For a two-wheeled robot (\(\Omega_s=2\)), one dimension is linear velocity and the other is yaw (angular velocity).

The robot constraints are managed using the Dynamic Window Approach (DWA), introduced by Fox et al. \cite{fox1997dynamic}. This technique sets a velocity window based on the current robot velocity and configuration. The window is then used to clip the robot's actions, as demonstrated in Algorithm \ref{alg:dwc}.

\begin{figure}[tp]
    \small
     \removelatexerror
    \begin{algorithm}[H]
    \caption{DynamicWindowClipping}\label{alg:dwc}
    \textbf{Inputs}: \\ $\- \- \-$ ${\mathrm{\acute a^r_t}}$: robot action as linear and angular velocity; ${v^r_t}$: current robot linear and angular velocity; Robot config\;
    $v_s \gets$ range of possible velocities based on minimal and maximal velocities of the robot\;     
    $v_d \gets$ range of velocities achievable in the next time step based on current velocity $v_r^t$ and by considering the minimal and maximal acceleration of the robot\;
    $v_{\cap} \gets$ intersection of $v_s$ and $v_d$\;
    ${\mathrm{\acute a^r_t}} \gets $  clip robot action ${\mathrm{\acute a^r_t}}$ with $v_{\cap}$\;
    \textbf{return} $\mathrm{\acute a^r_t}$\;
    \end{algorithm}
    \end{figure}

\subsection{Model Predictive Path Integral}\label{sec:mppi}

In MPPI \cite{williams2017information}, the objective function is based on the 'free energy' concept from control theory. This reformulates the expected return in \eqref{eq:reward} as:
\begin{equation}\label{eq:mppi_j}
\hat J(\pi^r_{gauss}, \pi^h_{nar}) = \log \mathbb{E}_{\tau}\left[\exp{\left(\frac{1}{\gamma}f_R(\tau)\right)} \right]\,,
\end{equation}
where \( \gamma > 0 \) is a scaling factor, or "temperature," influencing the exploration-exploitation trade-off. A higher $\gamma$ encourages exploration, while a lower $\gamma$ favors exploitation. The term \( \pi^h_{nar} \) represents the human policy, as introduced in Section \ref{sec:gcbc}.
"Free energy" refers to the system's usable energy (rewards) after accounting for entropy (uncertainty). This approach balances reward maximization with system adaptability, crucial for managing dynamic and uncertain environments.

\begin{figure*}[tp]
    \centering
    \begin{subfigure}[b]{1\textwidth}
        \centering
         \includegraphics[scale=0.6]{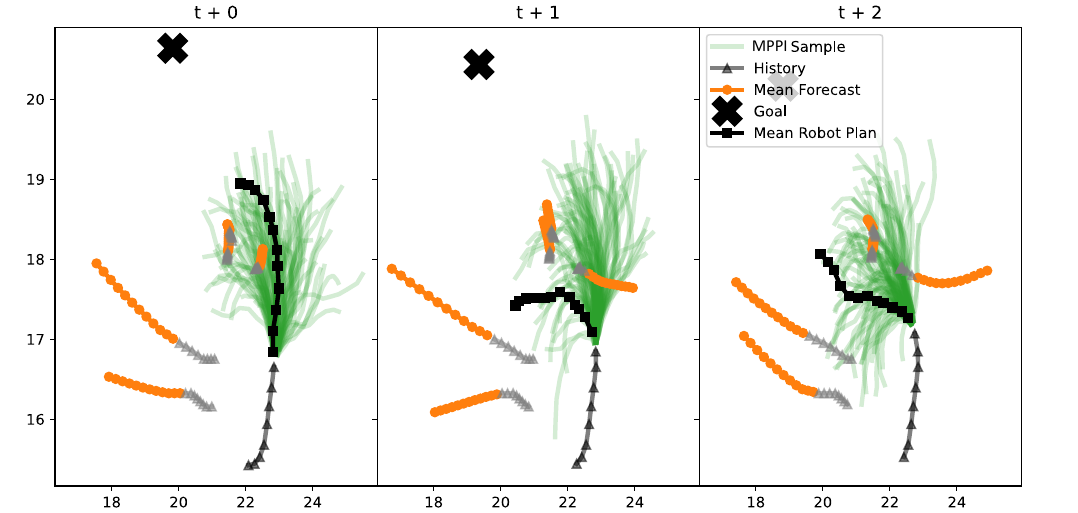}
    
    \end{subfigure}
    \\
    \begin{subfigure}[b]{1\textwidth}
        \centering
         \includegraphics[scale=0.607]{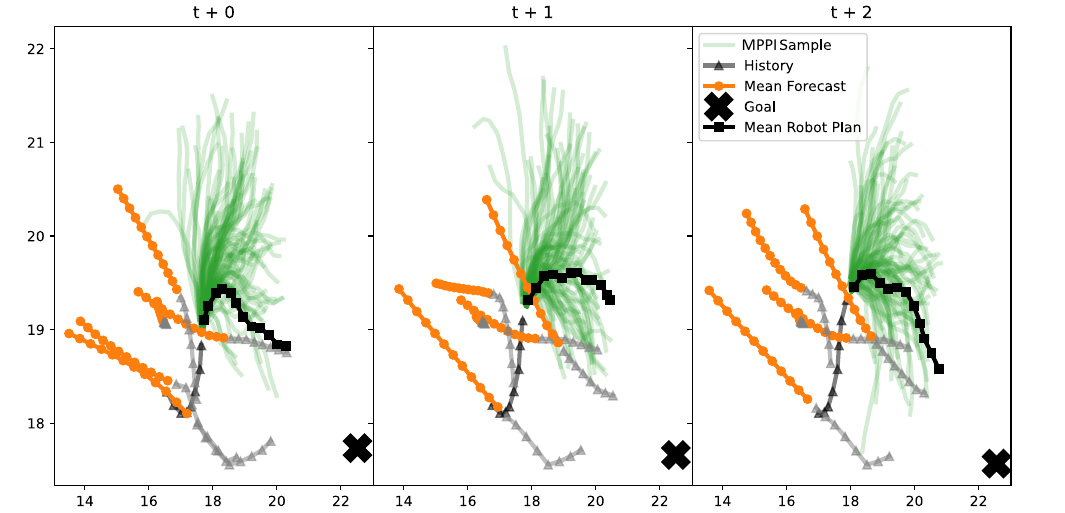}
 
    \end{subfigure}
    \caption{MPPI planning visualization for a two-wheeled robot (LoCoBot) \cite{locobotInterbotix} in the metric state space, with \( dt=0.4 \) s intervals. Columns show consecutive timesteps from two scenes, each depicted row-wise. Triangles indicate observed agent states. MPPI sample populations are green, with the mean best trajectory in black squares. The first state from this trajectory is executed. Human trajectory forecasts using the NAR model are shown as orange dots.
    }
      \label{fig:iterations}
   
\end{figure*}

Wagener et al. \cite{williams2017information} demonstrate that the optimization objective in \eqref{eq:mppi_j} can be refined by calculating the mean parameter of \( \pi^r_{gauss} \) through a weighted average. Each episode \( \tau \) from a batch of \( N_s \) episodes is assigned a weight:
\begin{equation}\label{eq:mppi_w}
\begin{aligned}	
\omega_n &= \text{softmax}_n\left(\frac{1}{\gamma}f_R(\tau_{1:N_s}) - \upsilon\right) \\
   &= \frac{e^{\frac{1}{\gamma}(f_R(\tau_{n}) - \upsilon)}}{\sum_{\grave n=1}^{N_s} e^{\frac{1}{\gamma}(f_R(\tau_{\grave n}) - \upsilon)}}\,.
\end{aligned}
\end{equation}
Here, \( \tau_{1:N_s} \coloneqq \{ \tau_{1}, \tau_{2}, \cdots, \tau_{N_s} \} \) represents the batch of trajectories, each corresponding to actions sampled from \( \pi^r_{gauss} \) and the human reactions from \( \pi^h_{nar} \). The term $\upsilon \coloneqq \max f_{R}(\tau_{1:N_s})$ ensures that at least one weight is non-zero, even if all returns are highly negative. Thus, trajectories with higher returns receive greater weights. The means \( \mu_{t} \) of the robot policy, guiding future actions, are updated accordingly:
\begin{equation}\label{eq:weighted_pert}
\mu_{t} =  \mu_{t} + \sum_{n=1}^{N_s} \omega_n \cdot (\tensor*[^n]{\mathrm {\acute a}}{^{r}_{t}} -  \mu_{t}),
\end{equation}
where \( \tensor*[^n]{\acute a}{^{r}_{t}} \) represents the action of the \( n \)-th episode at time step \( t \). This optimization can be iterated multiple times to improve \( \mu \), but a single iteration of MPPI is sufficient for the operational efficacy of a two-wheeled robot in this study. The application of MPPI optimization for navigating a two-wheeled robot in a crowd is shown in Figure \ref{fig:iterations}.

In anticipation of subsequent MPPI optimizations, a "warm-starting" process is implemented, a typical SMPC strategy. Instead of initializing \( \mu_{t} \) values to zero, warm-starting uses values from a previous SMPC optimization, \( \mu_{t}^{\text{old}} \), with a rolling operator. This operator initializes the policy \( \pi^r_{gauss} \) means from the previous solution, shifting by one time step: \( \mu_{t} \gets \mu_{t+1}^{\text{old}} \) for all states up to \( T-2 \), with the terminal mean at \( T-1 \) set to zero, as outlined in Algorithm \ref{alg:rolling_op}. This approach accelerates the optimization process, essential for real-time constraints requiring approximate solutions.
In summary, the MPPI algorithm determines the next robot action by introducing "white noise" to the best prior solution, simulating potential trajectories, computing the corresponding returns, and using a softmax function to calculate the weighted sum of these perturbed actions, as shown in Algorithm \ref{alg:mppi}. 

\begin{figure}[tp]
    \small
     \removelatexerror
    \begin{algorithm}[H]
    \caption{RollingOperator}\label{alg:rolling_op}
    \textbf{Inputs}: \\ $\- \- \-$ $\mu_{0:T-1}^{old}$: Robot policy $\pi^r_{gauss}$ means from previous SMPC optimization\;
    \For{$t=0$ \KwTo $T-2$ } {
        $\mu_t\gets \mu_{t+1}^{old}$\;
       }
    $\mu_{T-1}\gets$ initialize with zeros\;
    \textbf{return} $\mu_{0:T-1}$\;
    \end{algorithm}
\end{figure}

\begin{figure}[htp]
    \small
     \removelatexerror
    \begin{algorithm}[H]
    \caption{MPPI for Navigation in a Crowd}\label{alg:mppi}
    \textbf{Inputs}: \\ $\- \- \-$ $N_s$: number of samples; $\gamma\,$: temperature parameter; $\mathrm g^r\,$: robot goal position; $f_r, f_h$: state transition functions;
    $\pi^h_{nar}$: human policy; $dt$: duration of the timestep\;
    $\mu_t, \Sigma_t \gets (0, I)$ initialize parameters of robot policy $ \pi^r_{gauss}$ for every time step\;
        \While {\textit{task not completed}}{
             $\mathrm s_0,v^r_0 \gets $ observe the current states, as well as the linear and angular velocities of the robot\;
            \For{$n=1$ \KwTo $N_s$ } {
                $\mathrm g^h \gets $ forecast human goals as presented in Section \ref{sec:hgo}\;
                $\tensor*[^{n}]{\mathrm{\acute a}}{^{r}_{0:T-1}} \gets$ sample from $ \pi^r_{gauss}$ towards goal $\mathrm g^r$\;
                $R_n \gets 0$\;
                $\tensor*[^{n}]{v}{^{r}_{0}} \gets v^r_0$\;
                \For{$t=0$ \KwTo $T-1$ } {
                  $\tensor*[^{n}]{\mathrm{\acute a}}{^{r}_{t}} \gets$ DynamicWindowClipping($\tensor*[^{n}]{\mathrm{\acute a}}{^{r}_{t}}, \tensor*[^{n}]{v}{^{r}_{t}})$ \Comment{Algo. \ref{alg:dwc}}\;
                  $\tensor*[^{n}]{v}{^{r}_{t+1}}  \gets \tensor*[^{n}]{\mathrm{\acute a}}{^{r}_{t}}$\;
                  $\tensor*[^{n}]{ \mathrm a }{^{r}_{t}}  \gets \text{Pol2Cart}(\tensor*[^{n}]{\mathrm{\acute a}}{^{r}_{t}})\cdot dt$ \Comment{transform from polar to cartesian system}\;
                  $\mathrm a^h_{t} \gets$  sample from $ \pi^h_{nar}$ toward $\mathrm g^h$\;
                  ${R}_{n} \gets R_n + f_\phi(\tensor*[^{n}]{\mathrm s}{^{}_{t}}, \tensor*[^{n}]{\mathrm a}{^{r}_{t}}, \mathrm a^h_t)$\;
                   $\tensor*[^{n}]{\mathrm s}{^{r}_{t+1}} \gets f_r(\tensor*[^{n}]{\mathrm s}{^{r}_{t}}, \tensor*[^{n}]{\mathrm a}{^{r}_{t}})$\;
                  $ \tensor*[^{n}]{\mathrm s}{^{h}_{t+1}} \gets f_h(\tensor*[^{n}]{\mathrm s}{^{h}_{t}}, \tensor*[^{n}]{\mathrm a}{^{h}_{t}})$\;
                  $\tensor*[^{n}]{\mathrm s}{^{}_{t+1}} \gets $concatenate $\tensor*[^{n}]{\mathrm s}{^{r}_{t+1}} \oplus \tensor*[^{n}]{\mathrm s}{^{h}_{t+1}}$\;
                  }
            }
            $w_{1:N_s} \gets$ ComputeWeights(${R}_{1:N_s},\gamma$) \Comment{Algo. \ref{alg:comp_w}}\;
            \For{$t=0$ \KwTo $T-1$ } {
            $\mu_t\gets \bar \mu_t+\sum_{n=1}^{N_s} \omega_n \cdot (\tensor*[^n]{\mathrm{\acute a}}{^{r}_{t}}-\bar \mu_{t})$\;
           }
           SendToMotorController($\mu_0$)\;
           $\mu \gets$ RollingOperator($\mu^{}$) \Comment{init params. for next opimization, see Algo. \ref{alg:rolling_op}}\;

    }
    \end{algorithm}
\end{figure}

\begin{figure}[htp]
    \small
     \removelatexerror
    \begin{algorithm}[H]
    \caption{ComputeWeights}\label{alg:comp_w}
    \textbf{Inputs}: \\ $\- \- \-$ $f_R(\tau_{1:N_s})$: Sample Returns; $\gamma$: Temperature\;
            $\upsilon \gets \max f_R(\tau_{1:N_s})$\;
            \For{n=1 \KwTo $N_s$}{
                $\omega_n \gets softmax_n(\frac{1}{\gamma}(f_R(\tau_{1:N_s})-\upsilon))$\;
            }
    \textbf{return} $w_{1:N_s}$\;
    
    \end{algorithm}
\end{figure}

\subsection{Neural Autoregressive Model\label{sec:ar}}

In selecting a generative model for GCBC, we chose the Neural Autoregressive (NAR) model, based on insights from Moder et al. \cite{moder2023MIL}. This study shows that this model, when conditioned with goal prediction, performs comparably to models like GAN \cite{goodfellow2014generative} or VAE \cite{kingma2013auto}. The model’s ability to generate precise likelihood predictions aids in devising robot actions that replicate human behaviors. Advances in NLP, demonstrated by models like Llama2 \cite{touvron2023llama}, further underscore the NAR model's scalability.

The NAR model estimates the probability density of an agent’s actions:
\begin{equation}\label{eq:ar_h_policy}
    \pi^h_{\text{nar}} \coloneqq p(\mathrm A^{h}_t | \mathrm S_{\leq t}, G^{h}, \mathrm o; \theta_\text{nar}) = \mathcal{N}( \hat \mu_t^{h}, \hat \Sigma_t^{h}),
\end{equation}
with mean \(\hat \mu_t^{h}=f_{\mu}(\mathrm S_{\leq t},\mathrm G^{h}, \mathrm o; \theta_\text{nar})\) and covariance matrix \(\hat \Sigma_t^{h}=f_{\Sigma}(\mathrm S_{\leq t}, \mathrm G^{h}, \mathrm o; \theta_\text{nar})\), where \(\hat \mu_t^{h} \in \mathbb{R}^{\Omega_s} \) and \(\hat \Sigma_t^{h} \in \mathbb{R}^{\Omega_s \times \Omega_s}\). Functions \(f_\mu\) and \(f_\Sigma\) are neural networks with trainable parameters \(\theta_\text{nar}\), trained by maximizing the likelihood on the dataset \(\mathcal{D}_{\text{her}}\).
There is no distinction between the robot and humans; all predicted agents are assumed to be humans. The subsequent state is determined using the state transition function \(f_h\).

The NAR model uses an encoder-decoder architecture, as described by Moder et al. \cite{collos}. Each observation \(\mathrm o^k \) is processed through a transformer-encoder \cite{vaswani2017attention} to extract features. These features are then decoded using an LSTM \cite{HochSchm97}, with a Pooling Module (PM) to integrate the states of other agents into a unified feature vector. Training, as detailed in Section \ref{sec:gcbc}, maximizes the probability of human data relative to the specified goal. To promote collision-free states, a collision loss function is used, as detailed in \cite{collos}.

\subsection{Neural Inverse Autoregressive Model \label{sec:iar}}

Unlike the NAR model, the Neural Inverse Autoregressive (NIAR) model by Kingma et al. \cite{kingma2016improved} facilitates easier parallelization over time. The NIAR model's conditional probability distributions \( {p}(\mathrm A^{h}_t|\mathrm Z_{\leq t}^{h}, \mathrm G^h, \mathrm o; \theta_{\text{niar}}) \) are similar to those in \eqref{eq:ar_h_policy} but use \(\mathrm Z_{1:T}^{h} \overset{\text{iid}}{\sim} \mathcal{N}(0, I) \) for agent $h$, with $\mathrm Z_{\leq t}^{h} \coloneqq \mathrm Z_{1:t}^h$. These Gaussian conditional densities are parameterized by the mean $\check \mu_t^h = f_{\mu}(\mathrm Z_{\leq t}^{h}, \mathrm o; \theta_{\text{niar}})$ and covariance matrix $\check\Sigma_t^h = f_{\Sigma}(\mathrm Z_{\leq t}^h, \mathrm o; \theta_{\text{niar}})$, both computed using neural networks. Each conditional is independent of other agents and time steps, allowing future actions to be generated in parallel.
Each state is recursively determined using the transition function in \eqref{eq:fh}, starting from the initial observed state $\mathrm s_0^h$.

The NIAR model employs an encoder-decoder transformer architecture \cite{vaswani2017attention}. The encoder processes observed states \(\mathrm o^h \) and, with a PM, transforms them into a joint context vector. The decoder uses this context vector to predict all actions simultaneously, taking \(\mathrm Z^h \) and \(\mathrm G^h \) as inputs. The model uses causal attention, ensuring each action prediction is based on preceding inputs \(\mathrm Z^h_{\leq t} \), maintaining a sequential flow of information.

\subsection{Reward Function}\label{sec:reward}

The reward function is a sum of four distinct reward signals, represented as:
\begin{equation}
f_\phi(\mathrm A^r_t, \mathrm A^h_t, \mathrm S_t) = \sum_{i=1}^4 \lambda_i \phi_i \,,
\end{equation}
where each $\lambda_i$ is a weight parameter associated with the corresponding scalar reward signal $\phi_i$ and $\mathrm S_t$ represents here the concatenation $\mathrm S_t=\mathrm S^r_t \oplus \mathrm S^h_t$.

\subsubsection{The Reward Map}\label{sec:costmap_r}
As previously noted, a major challenge is the lack of comprehensive datasets that track human positions in real-world settings, along with contextual data crucial for robotic interpretation, such as occupancy grid maps. This gap means there's no data effectively integrating human behavior with the robot's environmental perspective.

To address this, a reward function incorporating a map is proposed for environments with static elements like walls and furniture, expressed as:
\begin{equation}\label{eq:costmap_r}
\phi_1 \coloneqq f_c(f_r(\mathrm A^r_t, \mathrm S^r_t)).
\end{equation}
The function \( f_c:\mathrm S_t^r \rightarrow \mathbb{R} \) assigns a real-valued reward to each state on a map, indicating the difficulty of navigating the environment (cost map). It guides the robot's assessment of navigability at a given position. For example, areas with obstacles receive lower rewards, signaling zones to avoid, while open areas get higher rewards, indicating safe navigation routes. By optimizing this reward function, the robot is encouraged to move towards high-reward areas and avoid low-reward ones, enhancing navigational efficiency and safety.

\subsubsection{Human Policy based Reward Signals}

The calculation of the following three reward signals is based on the human policy. First, the desired goal position for each human is identified (see Section \ref{sec:hgo}). The human policy \( \pi_{nar}^h \) from \eqref{eq:human_p} then predicts the actions of all \( k \) agents, including the robot, assuming humans perceive the robot as another human. As detailed in Section \ref{sec:gcbc}, the human policy and transition function enable autoregressive prediction of trajectories of length \( T \). The robot's plan is integrated by substituting the model-generated robot state with the state from the robot's pre-sampled plan, anticipating human reactions to the robot's plan.

The first reward signal is the collision-free reward. Plans where no collision occurs between the robot and humans at time step \( t \) are assigned a higher reward. The collision-free reward is defined based on the CoLoss as introduced by Moder et al. \cite{collos}, as:
\begin{equation}
     \phi_2 \coloneqq -\sum_{k=2}^K 1 - \operatorname{sig} (\beta(\|d_{t}^{rk}\|_2  - \gamma_{coll})),
\end{equation}
where \(\| d_{t}^{rk}\|_2  \coloneqq \| f_r(\mathrm A^r_t,\mathrm S^r_t) - f_h(\mathrm A^k_t, \mathrm S^k_t)\|_2 \) represents the Euclidean distance between the robot and the \( k \)-th agent. The sigmoid function is denoted by \( \operatorname{sig} \). The threshold \( \gamma_{coll} \) specifies the distance at which a collision is considered to occur, and \( \beta \) determines the precision of this discrimination.

Deriving an analytical reward for robot plans that imitate human behavior in complex scenarios is challenging. To address this, the human policy \( \pi_{nar}^h \) is used as a discriminator to evaluate how closely a plan resembles human behavior. The human-imitation reward is defined as\footnote{Here we neglect the sum over all dimensions $\Omega_s$ to avoid clutter.}:
\begin{equation}
    \phi_3 \coloneqq \log{\pi_{nar}^h}(\mathrm A^r_t|\mathrm S_{<t}^h, \mathrm S_{<t}^r, \mathrm g^r,\mathrm o; \theta_{nar}),
\end{equation}
where a high reward is given if a robot action \( a^r_t \) has a high log-likelihood. This reward function plays four critical roles in the SMPC algorithm design:

\begin{itemize}
    \item {Human Behavior Imitation}: This reward encourages the robot to imitate human behavior patterns, making its actions more understandable and predictable to nearby humans, thus fostering smoother interactions.
    
    \item {Ensuring Plans Remain within the Model's Distribution}: This reward aspect ensures the robot's plans stay within the model's predictions, avoiding "out of distribution" plans that could cause confusion or safety concerns in human-robot interactions.
    
    \item {Interface Between Human Policy and SMPC}: The human-imitation reward acts as an interface between the human policy and SMPC. It helps the robot use its understanding of human behavior to guide its actions, even if it cannot fully emulate human actions due to physical limitations.
    
    \item {Navigating in Environments with Static Obstacles}: Integrating environmental information into the human policy can enhance the robot's navigation capabilities in crowded settings with static obstacles. Essentially, the robot has the potential to learn from the intuitive navigation strategies humans employ around others and static objects, such as furniture or walls.
\end{itemize}

Inspired by the Social Influence Loss introduced by Moder et al. \cite{moder2022proactive}, this work designs a Social Influence Reward (SIR) to regulate how much the policy expects humans to avoid the robot. Expecting, too much space for the robot might lead to unsafe plans, while no clearance can cause the robot freezing problem. The SIR uses counterfactual reasoning to minimize the difference between conditioned and unconditioned predictions, based on a robot plan trajectory from SMPC. The SIR is defined as the summed Euclidean difference:
\begin{equation}
    \phi_4 \coloneqq -\sum_{k=2}^K \| \tensor*[]{\bar{\mathrm S}}{^{k}_{T}} - \tensor*[]{\mathrm S}{^{k}_{T}} \|_2,
\end{equation}
where $\mathrm{S}^{k}_{T}$ is the NAR prediction for agent $k$ at time $T$ conditioned on a robot trajectory, and $\mathrm{\bar S}^{k}_{T}$ is independent of the robot trajectory. Unlike the SI \cite{moder2022proactive}, the difference is considered only at the final time $T$ to avoid "punishing" robot plans that minimally impact humans' ability to reach their goals.

\subsection{Human Goal Optimization} \label{sec:hgo}

This paper discusses the use of goal-conditioned NAR or NIAR models to predict human movements based on predetermined goal positions $G^h$, but it does not yet detail the goal-setting method. Additionally, these models currently do not consider the environmental context, relying solely on the positional context of agents. This limitation can cause challenges in complex environments, like hospitals with many walls and narrow passageways, leading to impractical predictions, such as suggesting direct paths through walls, as shown in Figure \ref{fig:opti}.

Moder et al. \cite{moder2022proactive} introduce the Goal Flow model for forecasting endpoint goal positions. We propose here an alternative method to reduce computational time and the likelihood of human trajectories intersecting with static obstacles like walls. This method uses an NIAR model without goal conditioning. With this model, a batch of trajectories is predicted with size \(N_{s}\). The \(n\)-th action is denoted as \(\tensor*[^n]{\mathrm a}{^{k}_{t}}\). The trajectory with the highest reward map values \eqref{eq:costmap_r} and highest likelihood with respect to \({p}(\tensor*[^n]{\mathrm a}{^{k}_{t}}|{\mathrm Z}{^{k}_{<t}}, \mathrm o; \theta_{niar})\) is selected. The goal position for each individual is determined based on the last position in these optimal trajectories. This goal optimization method is summarized in Algorithm \ref{alg:hgo}.

\begin{figure}[tp]
    \small
     \removelatexerror
    \begin{algorithm}[H]
    \caption{ComputeHumanGoals}\label{alg:hgo}
    \textbf{Inputs}: \\ $\- \- \-$\( f_{{niar}} \): Trained NIAR model with weights \( \theta_{{niar}} \); $f_c$: Reward Map; $f_h$: Human transition function; $\mathrm o^h$: Observed human states \;
            \For{k=2 \KwTo $K$}{
                $R^k \gets$ Init with a very small number\;
                $n^\star \gets 1$ \;
                $\tensor*[^{n}]{ \mathrm{a}}{_{0:T-1}^{k}},\tensor*[^{n}]{ {\check \mu}}{_{0:T-1}^{k}}, 
                \tensor*[^{n}]{ {\check \Sigma}}{_{0:T-1}^{k}}  \gets $ 
                Forecast with NIAR and also store conditionals parameters\;
                \For{n=1 \KwTo $N_s$}{
                    $R^k_n \gets \sum_{t=0}^{T-1} \left[ f_c(\tensor*[^n]{\mathrm{s}}{^{h}_{t+1}}) \right. $\\
                    $\left. +  \log \mathcal{N}\left(\tensor*[^{n}]{\mathrm{a}}{_{t}^{k}}; \tensor*[^{n}]{ {\check \mu}}{_{t}^{k}}, 
                    \tensor*[^{n}]{ {\check \Sigma}}{_{t}^{k}}\right) \right]$\;
                    \If{$R^k_n > R^k$}{
                        $R^k \gets R^k_n $\;
                        $n^\star \gets n$\;
                    }
                }
                $\tensor*[^{}]{\mathrm g}{^{k}_{}} \gets \tensor*[^{n^\star}]{\mathrm s}{^{k}_{T}} $\Comment{get state traj. by applying $f_h$ to the actions}\;
            }
    \textbf{return} $\mathrm g^h$\;
    
    \end{algorithm}
\end{figure}
\begin{figure}[tp]
      \centering
      \includegraphics[scale=0.084]{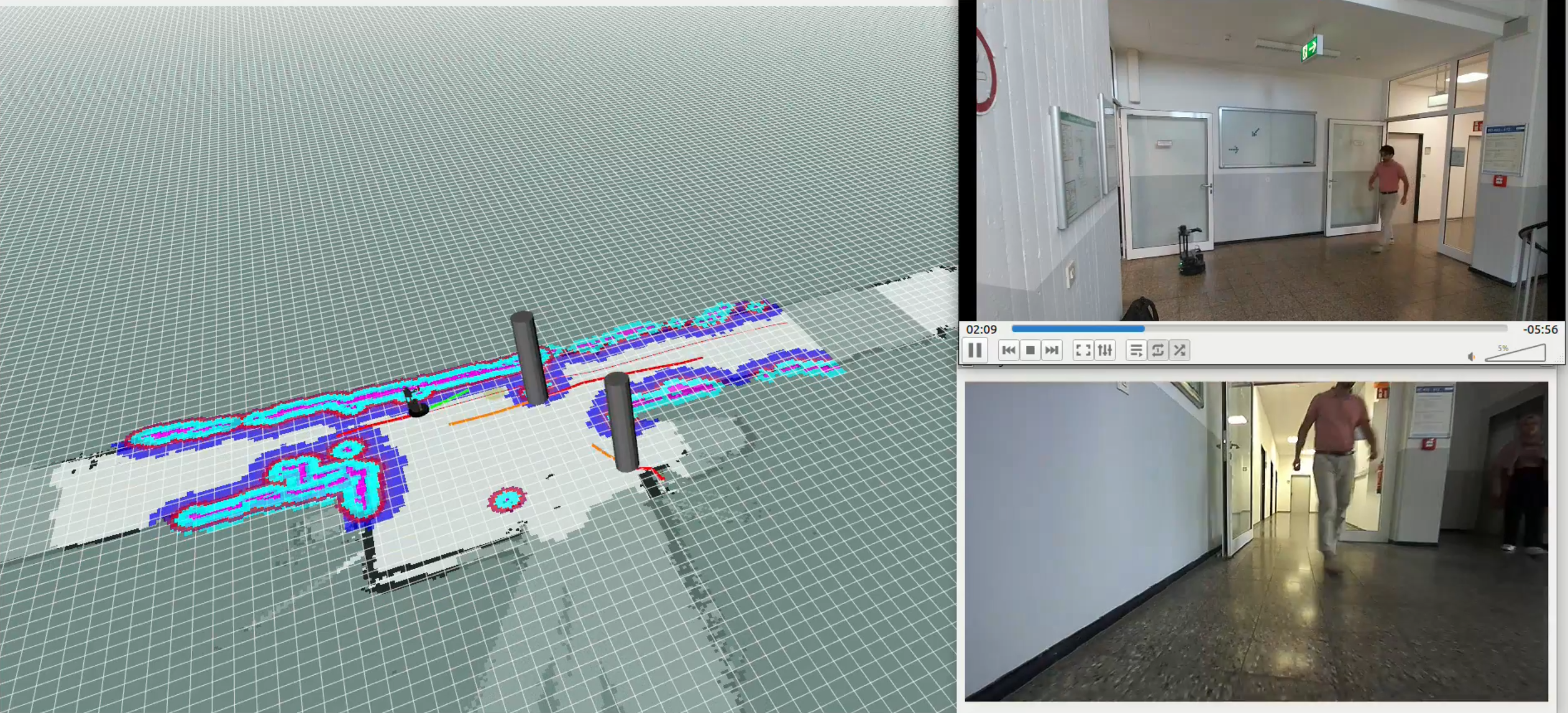}

      \caption{A snapshot from RViz shows humans as cylinders, the LoCoBot robot, and the environment as an occupancy grid map. Observed human states are in red, with the most likely predictions in orange. Human goals are chosen using Algorithm \ref{alg:hgo}.}
      \label{fig:opti}
\end{figure}

\subsubsection{Adaptive Sub-goal Navigation}
\label{subsection:subgoal}

With a map of the environment, our approach integrates seamlessly with a global planner, typically using a search-based algorithm like A-Star derived from the global occupancy map. Our method acts as a "local" planner, guiding the robot toward continuously updated sub-goals. Incorporating sub-goals from a global plan enhances navigation efficiency and avoids local minima, allowing the robot to interpret sensor data in the context of a broader strategy and identify optimal routes. For example, sub-goals help the controller recognize obstacles like walls in the global plan, facilitating efficient detours.

The velocity-adaptive sub-goal mechanism takes the global plan as input and outputs a sub-goal based on the robot's current velocity. This enables smoother velocity profiles, especially when navigating sharp turns in the global plan. The sub-goal dynamically adjusts to the robot's speed: higher velocities require a longer look-ahead distance for adapting to obstacles or turns, while slower speeds focus on immediate environmental details. The look-ahead range is constrained by the prediction horizon, robot dimensions, and reward map.

\section{Results}
We designed quantitative experiments using human datasets, as well as a real-world demonstration, with the intention of answering the following questions:
\textbf{Question 1}: How is the performance of our approach compared to a selected baseline?
\textbf{Question 2}: Can our algorithm outperform its individual components in collision avoidance and navigation tasks?
\textbf{Question 3}: How does our algorithm perform in the real world?

\subsection{Human Data Benchmark}
To address these queries, we evaluate our model and a baseline using real-world data instead of simulations, as real-world human interaction data better captures the complexity and unpredictability of human movements. Simulations often misrepresent these interactions, leading to an overestimation of algorithm performance.

\begin{figure}[tp]
    \centering
    \includegraphics[scale=0.35]{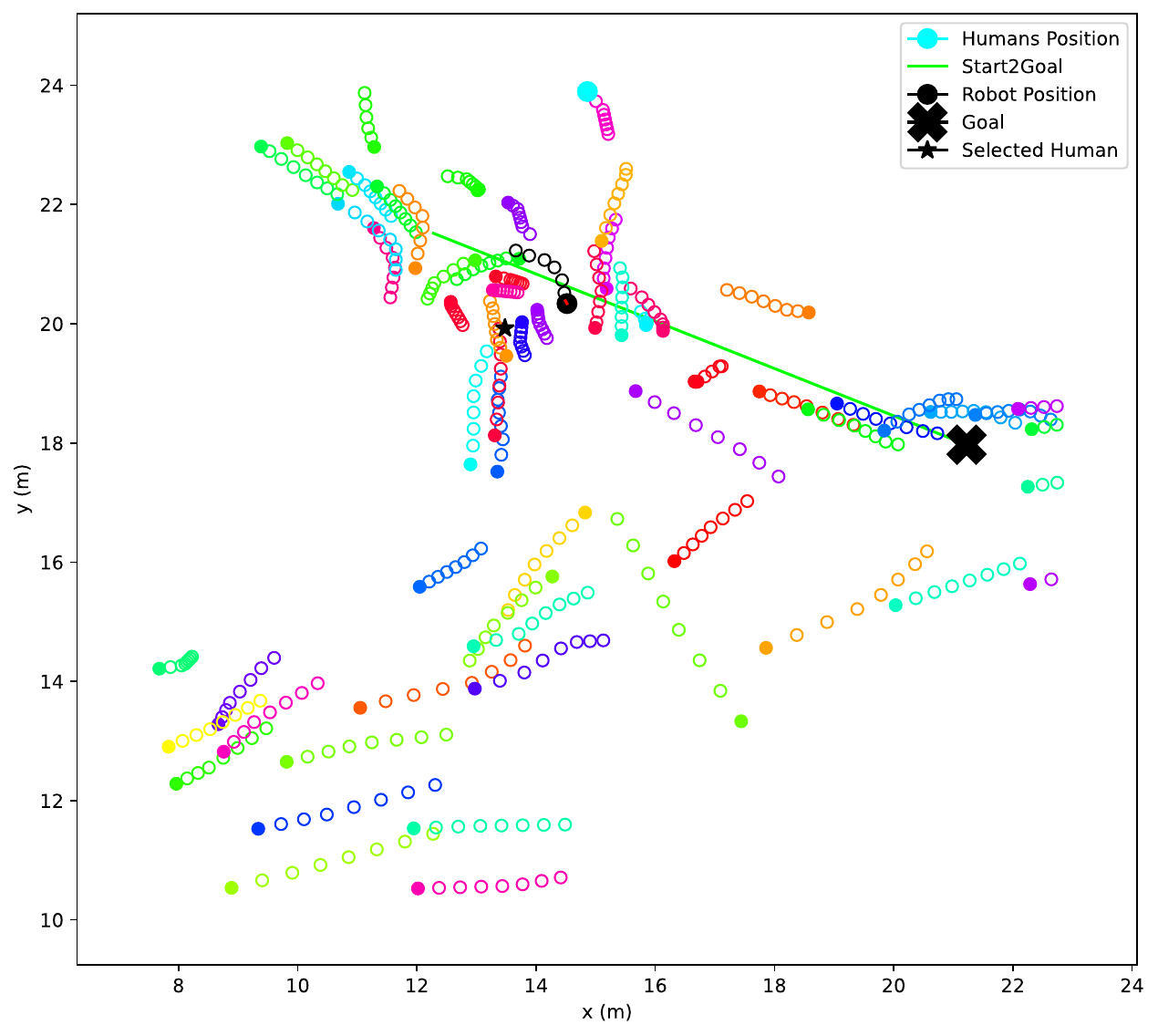}

    \caption{The robot (in black) navigates to its target while avoiding simulated humans (in various colors). Filled circles indicate the current position while unfilled circles demonstrate past positions.  The current position of the selected human, which is invisible to the robot, is marked with a star.}
    \label{fig:univ}
\end{figure}

We use the ETH \cite{pellegrini2010improving}, UCY \cite{leal2014learning}, L-CAS \cite{Sun20173DOFPT} and Wildtrack \cite{chavdarova} datasets. All data points are converted into world coordinates and interpolated at 0.4-second intervals. The joint dataset includes following subsets: two from ETH, three from UCY and one from Wildtrack and UCY respectively. For testing, we choose the most densely populated environment, "UNIV" from the UCY dataset, while the remaining datasets are used for training. The UNIV environment is divided into 412 individual scenes, each 20 seconds long (50 steps). The first 3.2 seconds (8 steps) of each scene, denoted as \( T_o \), are observed states. We evaluate robot navigation performance using the testing protocol of Moder et al. \cite{moder2023MIL}:

\begin{enumerate}
    \item Randomly select a human whose states are observable throughout the scene.
    \item Ensure the start and end positions of this human are at least $8$m apart; otherwise, choose a different human.
    \item Remove the selected human's states from the observation set after the first $T_o$ steps, so the robot cannot "see" them.
    \item Input the start and end positions of the selected human and the observed states of other agents into the navigation algorithm. Initialize the robot at the start position with the goal set to the end position of the selected human.
    \item If dynamics constraints are given, clip the robot's actions with DWC to ensure adherence to its dynamic constraints, regardless of the algorithm being evaluated.
\end{enumerate}
In Figure \ref{fig:univ}, an exemplary scene using the evaluation protocol and the introduced parameters is visualized.

This protocol provides a realistic benchmark for robot navigation in populated environments. The LoCoBot \cite{locobotInterbotix} is chosen as a representative platform, characterized by a linear speed of 0.7 m/s, angular speed of 1.0 rad/s, linear acceleration of 0.5 m/s², and angular acceleration of 3.2 rad/s². Each approach is evaluated 10 times due to the random human selection, with results summarized as mean values.

In the UNIV dataset, the average population density around the selected human within a 3 m radius is 0.3 humans/m². Each goal position, typically 10.2 m away from the start, is feasibly reachable by a human. The NAR and NIAR models predict the next 12 timesteps, with the predicted goal position being, on average, 3.2 m away. The human trajectory in each scene provides insights into potential human behaviors, offering a meaningful benchmark for comparison.

\subsubsection{Evaluation Metrics}
The metrics are defined as follows:
\begin{itemize}
    \item \textbf{Success}: Percentage of robots that reach their goal without colliding.
    \item \textbf{Coll$_{<21}$}: Percentage of robots colliding with humans when the distance to their center points is below 0.21 m.
    \item \textbf{Coll$_{<31}$}: Percentage of robots within 0.31 m of human center points.
    \item \textbf{Timeout}: Percentage of robots that fail to reach the goal within 16.4 s plus an additional 8 s tolerance.
    \item \textbf{Freezing Behavior (FB)}: Percentage of robot paths that are 1.25 times longer than the corresponding human path.
    \item \textbf{Max Freezing Behavior (maxFB)}: Highest ratio of robot path length to the corresponding human path length, expressed as a percentage.
\end{itemize}

\subsubsection{Baselines}

The MPPI approach, integrated with the NAR forecasting model, is designated as \textbf{MPPI-NAR}. The following algorithms are selected as baselines:

\begin{itemize}
    \item \textbf{DWA} \cite{fox1997dynamic}: A sampling-based navigation algorithm widely used in ROS, serving as a practical benchmark.
    \item \textbf{DWA-NAR}: Integrates DWA with the goal-conditioned NAR model. Goals are determined using the methodology in Section \ref{sec:hgo}.
    \item \textbf{GCBC-NAR} and \textbf{GCBC-NIAR}: Variants of Goal-Conditioned Behavioral Cloning using NAR and NIAR models, respectively. Goals are established through human goal optimization in Section \ref{sec:hgo}. Unlike MPPI and CEM, these models execute only the next most probable action without extensive planning. It is noteworthy that the GCBC-NIAR serves as a baseline for a robot that only sees the goal and not humans.
    \item \textbf{CEM-Hybrid} \cite{moder2023MIL}: Employs the Cross Entropy Method (CEM) for SMPC, using a hybrid NIAR and NAR goal-conditioned model with DWC. It optimizes in the latent space of the NIAR model, conducting stochastic optimization over three iterations, compared to one in MPPI.
    \item \textbf{CQL} \cite{kumar2020conservative}: An offline RL approach using $\mathcal{D}_{\text{her}}$ to learn a conservative Q-function based on expert actions, mitigating the risk of overestimating states not in the expert dataset. MPPI-NAR actions are used as substitute expert data.
    \item \textbf{TD3+BC} \cite{fujimoto2021minimalist}: An offline RL method adding a behavior cloning term to policy updates and normalizing data, achieving comparable performance to CQL with reduced computational overhead. MPPI-NAR actions are used as expert data.
    \item \textbf{TD3} \cite{fujimoto2018addressing}: Is selected for the evaluation of online Actor-Critic RL. The testing protocol is adapted to an OpenAI Gym \cite{brockman2016openai} environment to facilitate online training. Within this setting, individual policies engage in exploration while considering robot dynamics, in scenarios where simulated humans act as if unaware of the robot's presence. The strategy of pre-training TD3 Actor and Critic components using TD3+BC.
\end{itemize}

It is noted that the training dataset has fewer interactions than the test dataset. Due to unsatisfactory initial results from RL models trained on the training dataset, CQL, TD3, and TD3+BC are allowed to be trained on the test dataset. Further details on the implementation are provided in Section \ref{sec:impl_de}.


\begin{table*}[t!h]
    \caption{\label{tab:quan} {UNIV benchmark results for human crowd navigation were computed on an NVIDIA 2080Ti graphics card. Each approach was evaluated 10 times (due to the random selection of the human) and summarized as mean values. The algorithms were tested with LoCoBot dynamics constraints. Metrics are detailed in the main text. }}

    \centering
    \begin{tabular}{lrrrrrrr}
        \hline
        Algorithm                                                                       & \multicolumn{1}{l}{\begin{tabular}[c]{@{}l@{}}Success \\ in \%\end{tabular}} & \multicolumn{1}{l}{\begin{tabular}[c]{@{}l@{}}Coll$_{< 21}$\\ in \%\end{tabular}} & \multicolumn{1}{l}{\begin{tabular}[c]{@{}l@{}}Coll$_{< 31}$ \\ in \%\end{tabular}} & \multicolumn{1}{l}{\begin{tabular}[c]{@{}l@{}}Timeout \\ in \%\end{tabular}} & \multicolumn{1}{l}{\begin{tabular}[c]{@{}l@{}}FB\\ in \%\end{tabular}} & \multicolumn{1}{l}{\begin{tabular}[c]{@{}l@{}}maxFB\\ in \%\end{tabular}} & \multicolumn{1}{l}{\begin{tabular}[c]{@{}l@{}}Runtime \\ in ms\end{tabular}} \\ \hline
        \multicolumn{8}{c}{(a) Comparison with Baselines}                                                                                                                                                                                                                                                                                                                                                                                                                                                                                                                                                                                                          \\ \hline
        \multicolumn{1}{l|}{DWA}                                                        & 38.5                                                                         & 56.0                                                                              & 75.8                                                                               & 5.5                                                                          & 4.0                                                                    & 170                                                                       & 47.1                                                                         \\
        \rowcolor[HTML]{EFEFEF} 
        \multicolumn{1}{l|}{\cellcolor[HTML]{EFEFEF}DWA-NAR}                            & 71.1                                                                         & 20.4                                                                              & 44.3                                                                               & 8.5                                                                          & 1.7                                                                    & 154                                                                       & 42.6                                                                         \\
        \multicolumn{1}{l|}{GCBC-NAR}                                                   & 40.3                                                                         & 59.4                                                                              & 84.0                                                                               & 0.3                                                                          & 0                                                                      & 104                                                                       & 15.2                                                                         \\
        \rowcolor[HTML]{EFEFEF} 
        \multicolumn{1}{l|}{\cellcolor[HTML]{EFEFEF}GCBC-NIAR}                          & 47.7                                                                         & 51.1                                                                              & 80.6                                                                               & 1.1                                                                          & 0                                                                      & 102                                                                       & 9.0                                                                          \\
        \multicolumn{1}{l|}{CQL}                                                        & 15.0                                                                         & 70.7                                                                              & 87.8                                                                               & 14.4                                                                         & 51.7                                                                   & 211                                                                       & 2.0                                                                          \\
        \rowcolor[HTML]{EFEFEF} 
        \multicolumn{1}{l|}{\cellcolor[HTML]{EFEFEF}TD3}                                & 78.1                                                                         & 21.3                                                                              & 45.1                                                                               & 0.6                                                                          & 2.8                                                                    & 163                                                                       & 2.4                                                                          \\
        \multicolumn{1}{l|}{TD3+BC}                                                     & 26.1                                                                         & 70.0                                                                              & 88.2                                                                               & 0.4                                                                          & 19.8                                                                   & 218                                                                       & 1.9                                                                          \\
        \rowcolor[HTML]{EFEFEF} 
        \multicolumn{1}{l|}{\cellcolor[HTML]{EFEFEF}CEM-Hybrid}                         & 91.2                                                                         & 8.6                                                                               & 32.1                                                                               & 0.0                                                                          & 0.7                                                                    & 136                                                                       & 35.5                                                                         \\
        \multicolumn{1}{l|}{MPPI-NAR}                                                   & 89.4                                                                         & 10.4                                                                              & 39.5                                                                               & 0.2                                                                          & 0.9                                                                    & 144                                                                       & 15.8                                                                         \\ \hline
        \multicolumn{8}{c}{(b) Ablation Study}                                                                                                                                                                                                                                                                                                                                                                                                                                                                                                                                                                                                                     \\ \hline
        \multicolumn{1}{l|}{CEM-NAR}                                                    & 89.2                                                                         & 10.7                                                                              & 39.5                                                                               & 0.1                                                                          & 0.2                                                                    & 126                                                                       & 41.0                                                                         \\
        \rowcolor[HTML]{EFEFEF} 
        \multicolumn{1}{l|}{\cellcolor[HTML]{EFEFEF}CEM-NIAR}                           & 91.8                                                                         & 7.5                                                                               & 26.2                                                                               & 0.7                                                                          & 0.5                                                                    & 152                                                                       & 19.1                                                                         \\
        \multicolumn{1}{l|}{CEM-Hybrid-L}                                               & 87.0                                                                         & 11.4                                                                              & 37.5                                                                               & 1.7                                                                          & 7.4                                                                    & 200                                                                       & 35.1                                                                         \\
        \rowcolor[HTML]{EFEFEF} 
        \multicolumn{1}{l|}{\cellcolor[HTML]{EFEFEF}CEM-Hybrid-ST}                      & 90.6                                                                         & 9.3                                                                               & 34.1                                                                               & 0.2                                                                          & 0.5                                                                    & 143                                                                       & 35.1                                                                         \\
        \multicolumn{1}{l|}{MPPI-NIAR}                                                  & 92.7                                                                         & 7.1                                                                               & 28.3                                                                               & 0.2                                                                          & 1.3                                                                    & 163                                                                       & 9.4                                                                          \\
        \rowcolor[HTML]{EFEFEF} 
        \multicolumn{1}{l|}{\cellcolor[HTML]{EFEFEF}{\color[HTML]{000000} MPPI-NAR-ST}} & {\color[HTML]{000000} 82.9}                                                  & {\color[HTML]{000000} 17.1}                                                       & {\color[HTML]{000000} 50.5}                                                        & {\color[HTML]{000000} 0.1}                                                   & {\color[HTML]{000000} 0.7}                                             & {\color[HTML]{000000} 148}                                                & {\color[HTML]{000000} 15.6}                                                  \\
        \multicolumn{1}{l|}{MPPI-NAR-NSI}                                               & 88.5                                                                         & 11.3                                                                              & 39.7                                                                               & 0.2                                                                          & 1.3                                                                    & 157                                                                       & 15.3                                                                         \\ \hline
        \end{tabular}

\end{table*}

\subsubsection{Benchmark Results and Discussion}
\textbf{Towards Question 1.} 
The test protocol, which includes scenarios enforcing robot dynamic constraints (step 5), evaluates the algorithms. The data in Table \ref{tab:quan}(a) show that SMPC approaches MPPI-NAR and CEM-Hybrid outperform the baseline in success rate. Compared to MPPI-NAR, CEM-Hybrid is slightly superior in providing more efficient collision avoidance, although its runtime is observed to be 225\% longer. This increase in computation time is attributed to the requirement of three optimization iterations for CEM, whereas MPPI does not benefit from additional iterations. Additionally, the DWA-NAR results indicate that sample-based planning proves more effective when planning for every future action rather than focusing solely on the next action in relation to the current state.

The importance of covariate shift is highlighted by the “offline” approaches, GCBC and CQL. These models can guide the robot towards the goal but fail to fully learn the collision avoidance policy for a two-wheeled robot from data alone. In contrast, TD3 performs better, emphasizing the importance of online RL in addressing covariate shift. However, online RL results are imperfect despite various hyperparameters and training techniques. Similar challenges in deploying RL for navigation in crowded environments are noted in the literature (e.g., \cite{moder2023MIL,trautman2020real}), supporting these findings and suggesting further refinement of RL methods is needed. Additionally, it is noted that RL methods are highly sensitive to changes in their environment. When trained on a training dataset that features scenarios less crowded than those in the "Univ" test dataset, lower performance is observed. In contrast, higher results are achieved when training occurs directly on the test dataset (which are presented in Table \ref{tab:quan}(a)), characterized by more crowded and complex situations.

Surprisingly, in this context, the GCBC-NIAR model outperforms the GCBC-NAR model, even though the NAR model is designed to incorporate interactions with other humans. We argue that the NAR model often encounters unknown states and makes errors, particularly because it actively avoids collisions. This underscores the importance of the search algorithm. Moreover, both the NAR-GCBC and NIAR-GCBC models outperform CQL and TD3+BC, highlighting the importance of goal-seeking behavior in this benchmark. Despite these challenges, there is optimism that with additional data and in environments featuring more interaction with other agents, the robustness and informativity of the NAR model will improve.

\textbf{Towards Question 2.} 
An ablation study is conducted for the proposed SMPC algorithms. It examines the impact of different models: CEM-NAR uses only the NAR model, and CEM-NIAR uses only the NIAR model for optimization. The study also explores the effects of optimizing with stochastic forecasts of human movements, labeled as CEM-Hybrid-ST and MPPI-NAR-ST. Additionally, it investigates optimization in the latent space of the Hybrid Model, with results labeled as CEM-Hybrid-L. Finally, MPPI-NAR-NSI shows optimization without the SIR. The results are presented in Table 1(b).

Table \ref{tab:quan}(b) demonstrates that the NIAR model is the fastest and most effective, as expected. It assumes humans are unaffected by any agent, with the test environment designed for humans to ignore the robot. In contrast, the NAR model performs poorly in collision avoidance but excels in FB, as it expects humans to yield space to the robot, helping to prevent the robot freezing problem in real-world scenarios. To avoid the robot “expecting” too much space, the SIR can regulate this behavior, improving performance as seen in the comparison between MPPI-NAR and MPPI-NAR-NSI. The hybrid model shows promise in balancing the flexibility of the NAR and NIAR models, suggesting an optimal plan might be achievable even without a direct SIR.

In comparison, MPPI consistently outperforms CEM, despite MPPI requiring only one iteration. CEM, with its multiple optimization iterations, may be better suited for stochastic predictions. Optimization in the hybrid model’s latent space with CEM does not yield expected results, possibly due to the robot’s movements deviating too much from human movements, resulting in the highest maxFB rate. These findings highlight the importance of understanding human movement. Policies that accurately capture these dynamics improve success, FB, and maxFB outcomes, while overly rigid policies, like moving straight ahead, quickly reach their limits.

\begin{figure*}[!htp]
    \centering
    \begin{subfigure}{.3\textwidth}
        \centering
        \includegraphics[scale=0.23]{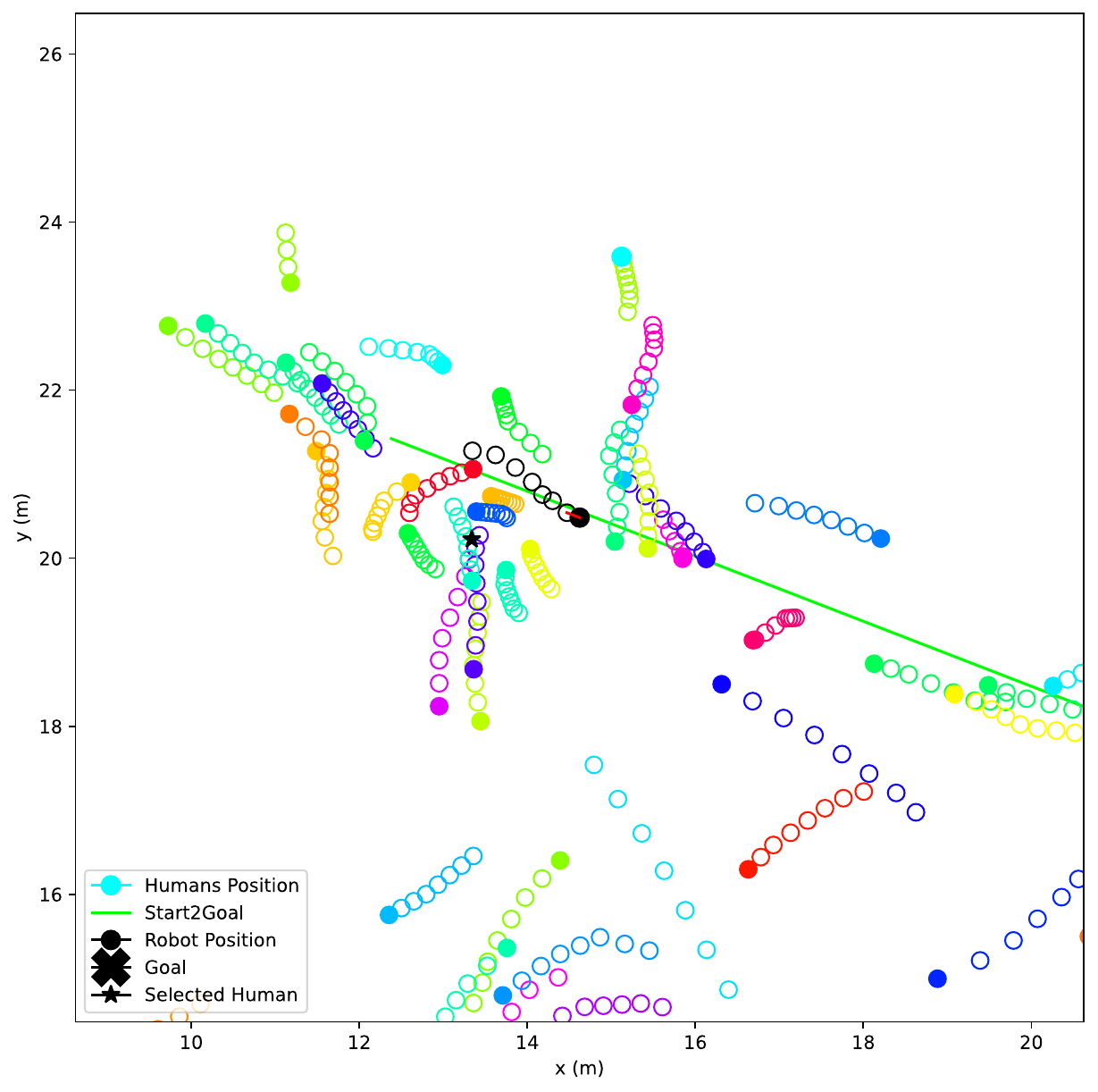}
\caption{ }
    \end{subfigure}
    \begin{subfigure}{.3\textwidth}
        \centering
        \includegraphics[scale=0.23]{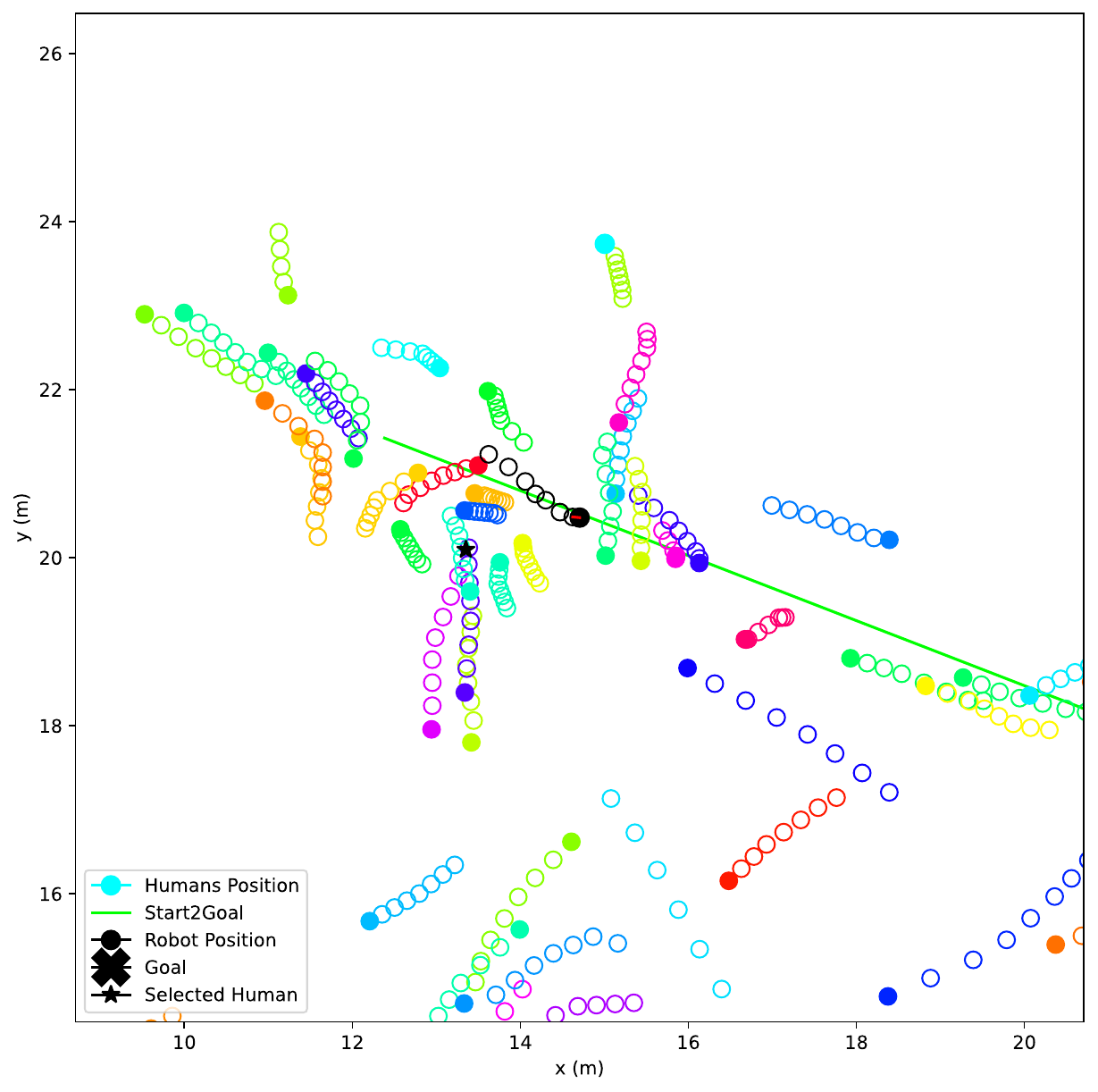}
\caption{ }
    \end{subfigure}
    \begin{subfigure}{.3\textwidth}
        \centering
       \includegraphics[scale=0.23]{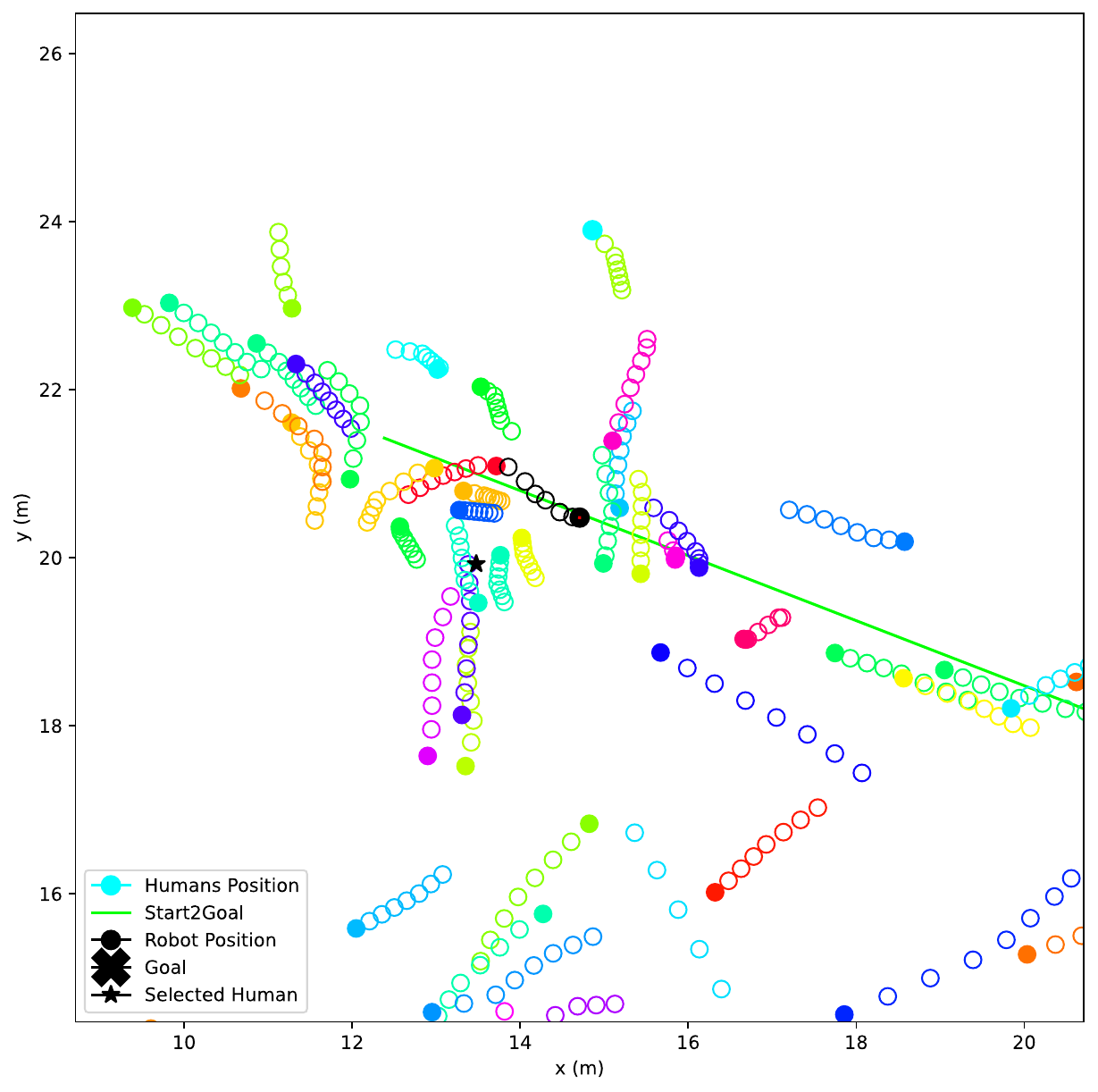}
\caption{ }
    \end{subfigure}
    
    \vspace{0.3cm}
    
    \begin{subfigure}{.3\textwidth}
        \centering
        \includegraphics[scale=0.23]{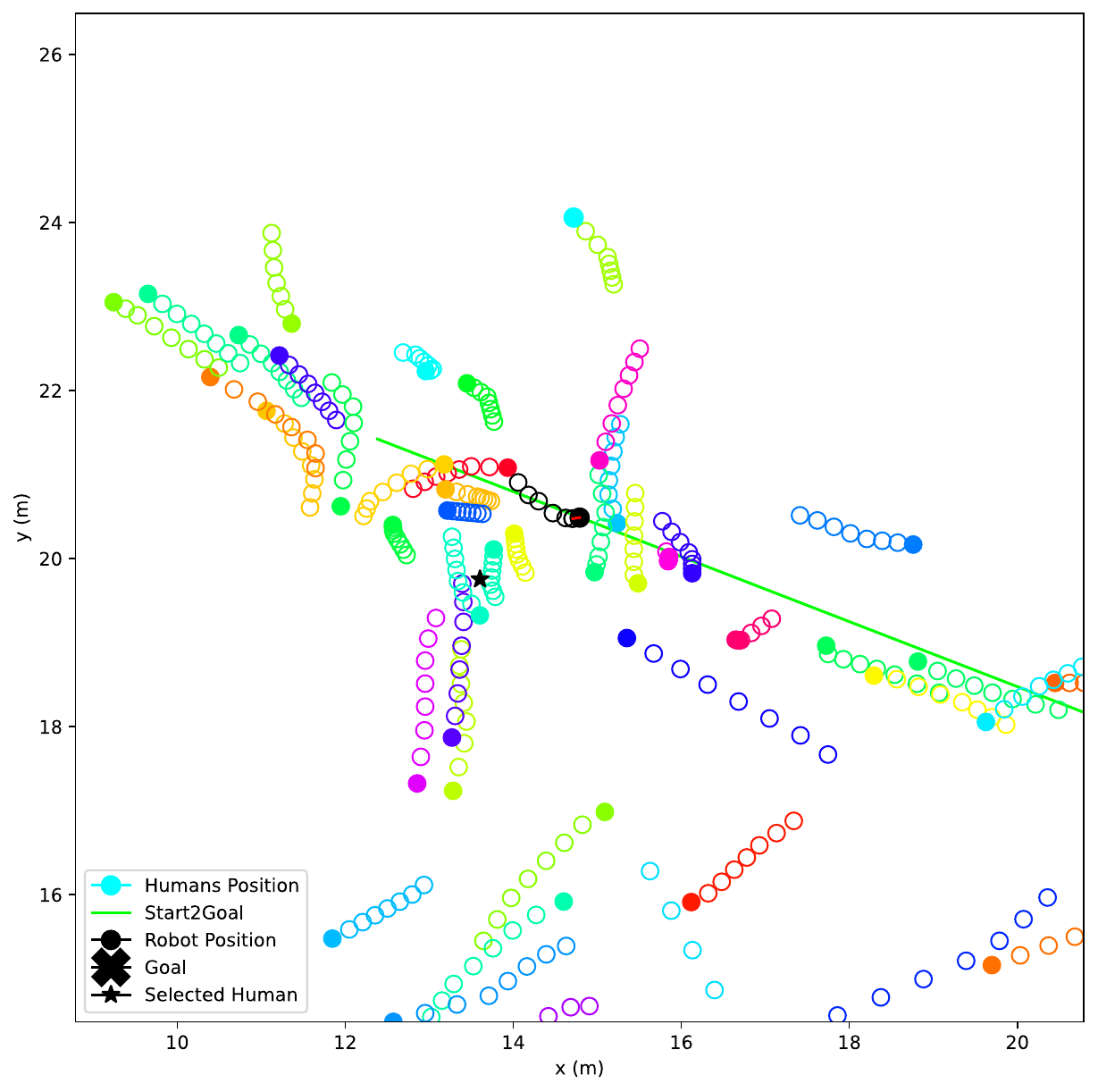}
\caption{ }
    \end{subfigure}
    \begin{subfigure}{.3\textwidth}
        \centering
       \includegraphics[scale=0.23]{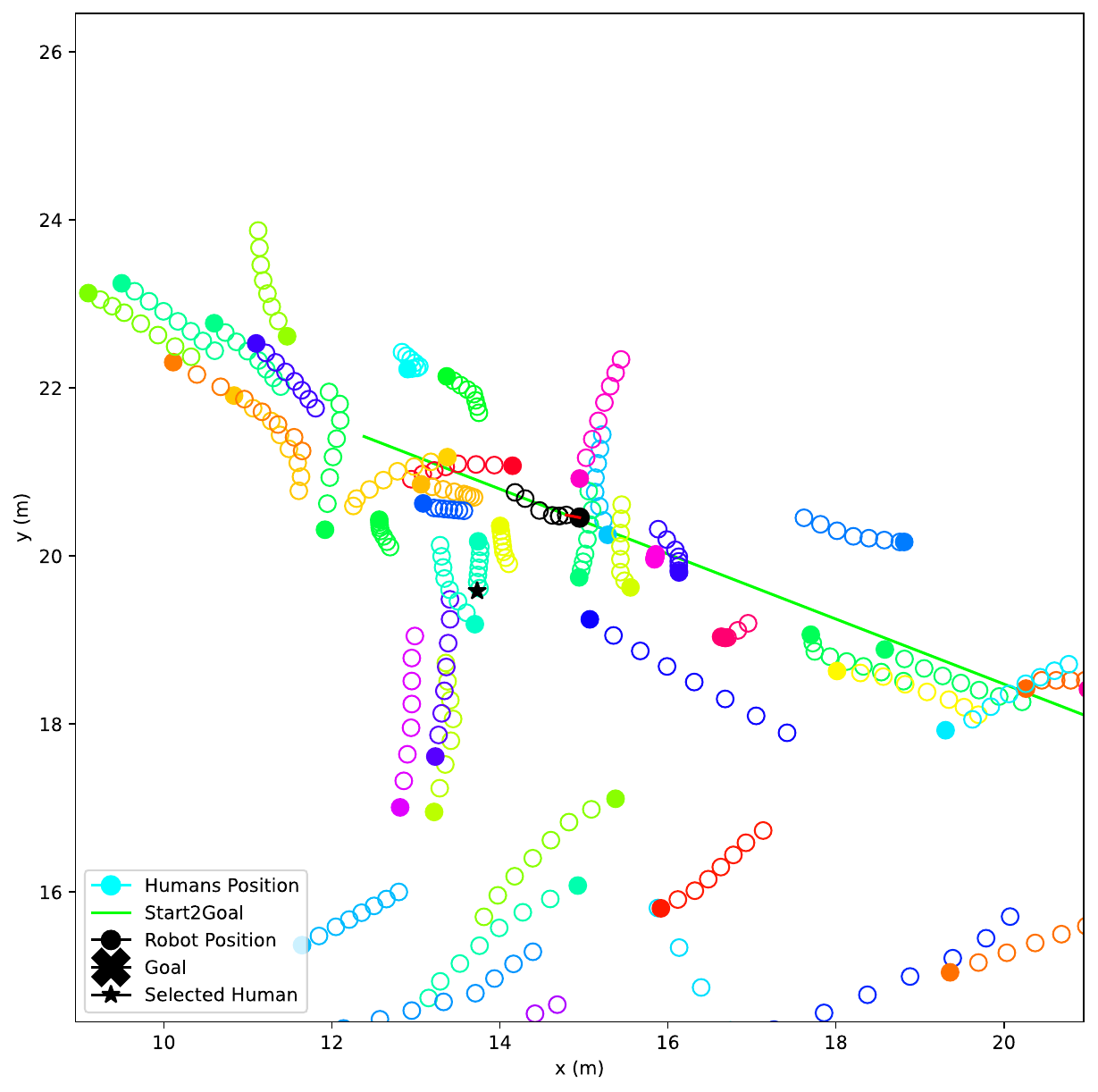}
        \caption{ }
    \end{subfigure}
    \begin{subfigure}{.3\textwidth}
        \centering
       \includegraphics[scale=0.23]{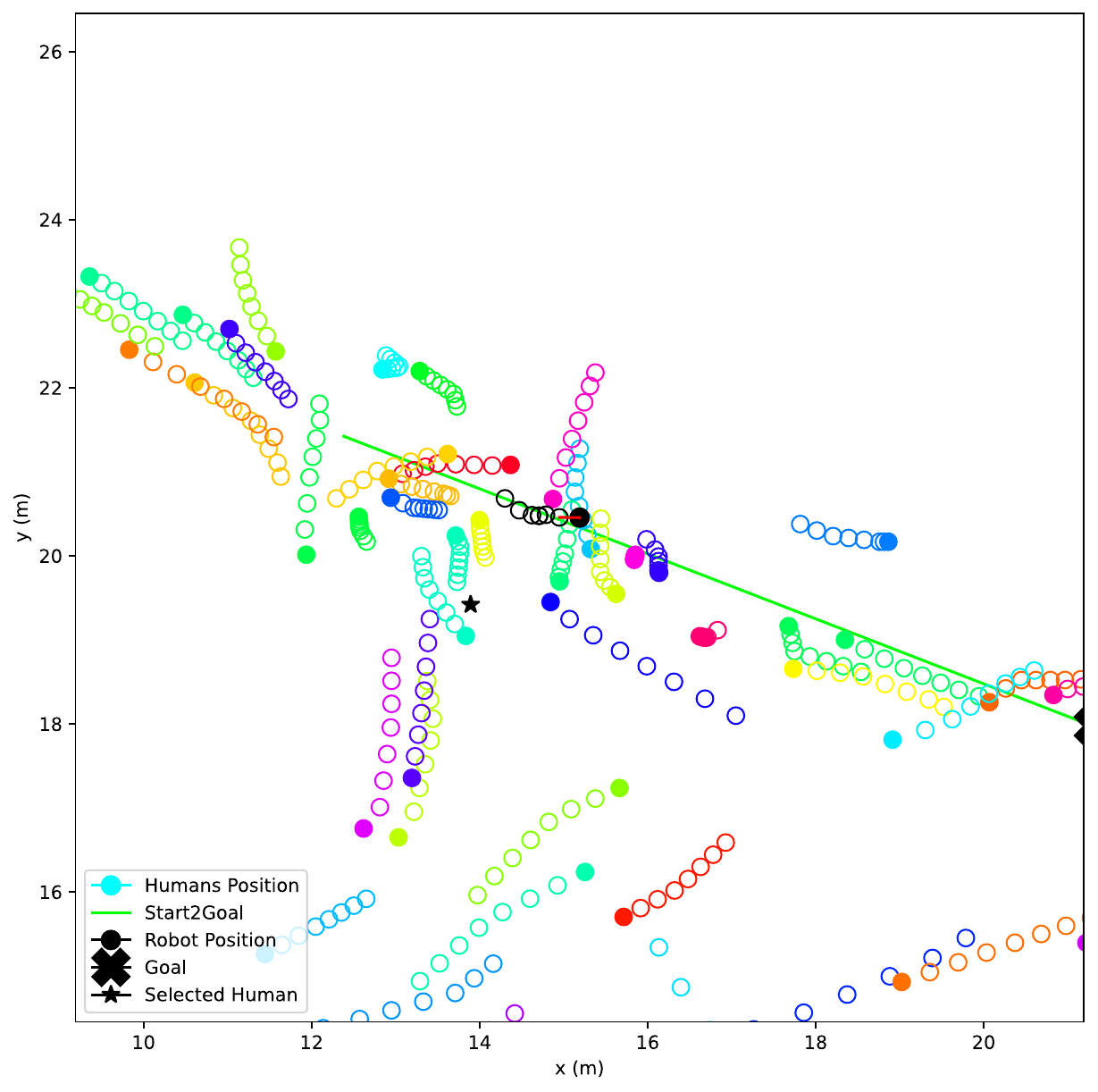}
        \caption{ }
    \end{subfigure}
    
    \caption{The figure shows a sequence (a-f) where the robot (black) navigates to its target while avoiding simulated humans (various colors). The robot stops to let crossing humans pass, then accelerates towards its target, avoiding collisions.}
    \label{fig:quanti-eval}
\end{figure*}

A qualitative evaluation, shown in Figure \ref{fig:quanti-eval}, demonstrates that the MPPI-NAR algorithm not only achieves basic collision avoidance but also engages in proactive planning, such as the robot pausing to let a person pass before proceeding. More videos and visualizations can be found \href{https://human-maze-navigation.github.io/}{here}, including a demonstration of what happens when SIR has a negative value, causing the robot to intentionally interrupt people.

\subsubsection{Implementation Details}\label{sec:impl_de}

Notably, the ETH\&UCY datasets do not label the physical shape of humans, so a collision event is determined based on Euclidean distances, with less than 0.2m defined as a collision, following previous works \cite{collos, moder2022proactive, trautman2015robot, moder2023MIL}. Accordingly, the collision cost distance parameter, $\gamma_{coll}$, is set to 0.2m. For both training and inference, 8 states are observed ($T_o \coloneqq 8$) and 12 states are predicted ($T \coloneqq 12$). Shorter history sequences are padded with zeros. To increase efficiency, only the five humans closest to the robot within a 5m radius are considered. The goal and all states are represented relative to the robot’s current position, reducing reliance on fixed world coordinates. To ensure consistency and prevent encountering unseen values, the goal distance is capped at 10m and limited to a minimum of 3m to maintain a consistent speed as the robot moves toward endpoints rather than waypoints.

The following parameters are determined empirically: the weighting factor for avoiding collisions with humans, $\lambda_2$, is set to $10^3$; the weighting factors for avoiding collisions with the environment, $\lambda_1$, and for human behavior imitation, $\lambda_3$, are both set to 1; $N_s:=800$ samples are taken per iteration. Additionally, the temperature parameter $\gamma$ is set to 1, $\beta$ is set to 35, and $\gamma_{coll}$ is set to $0.2$m. For CEM, the default number of iterations is set to three.

The reward function for CQL, TD3, and TD3+BC differs from that discussed in Section \ref{sec:reward}. This function awards a positive reward for approaching the goal and applies a negative reward for collisions or timeouts, tailored to the robot’s proximity to the goal and nearby humans. The d3rlpy framework \cite{d3rlpy} is used for the RL implementation, and PyTorch is used for the neural network implementation. For further details and to access the dataset used, interested readers are referred to the following repository: \href{https://human-maze-navigation.github.io/}{code}.

\subsection{Real-World Demonstrations}\label{sec:rl_demo}
\begin{figure}[ht]
      \centering
        \includegraphics[scale=0.4]{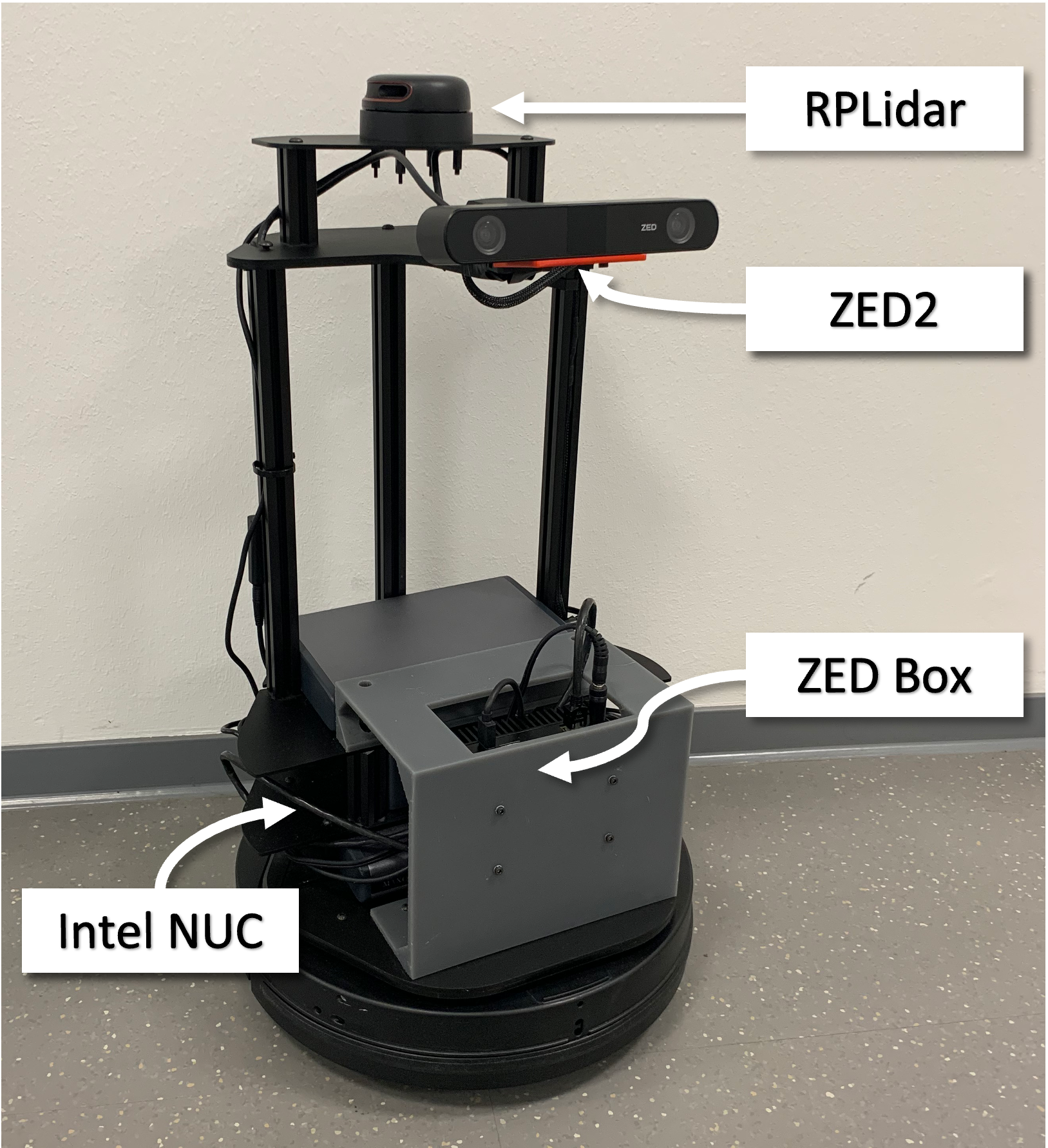}

        \caption{The Locobot’s sensory system includes a Stereolabs ZED 2 stereo camera and a 2D RPLidar A2M8 laser scanner. The Intel NUC handles localization, mapping, and navigation tasks, while the ZedBox manages human detection and tracking using the ZED 2 camera.}
      \label{fig:locobot}
\end{figure}

\textbf{Towards Question 3.} To accomplish real-world locomotion tasks, the MPPI-NIAR\footnote{Ideally, the MPPI-NAR model would be the subject of testing; however, due to hardware limitations, only the NIAR model can be evaluated without experiencing substantial latency.} algorithm is implemented on the mobile robot platform LoCoBot \cite{locobotInterbotix}, as depicted in Figure \ref{fig:locobot}, using ROS 2 Humble. This open-source, differential-drive robot is equipped with a 2D lidar and an Intel NUC featuring an 8th Gen Intel Core i3 processor. The system is upgraded by replacing the default Intel Realsense camera with the more advanced Stereolabs ZED 2 3D camera and augmenting it with an additional computing unit, the ZedBox, powered by an Nvidia Jetson Xavier NX board.

Localization, mapping, and navigation are managed by the Intel NUC, running SLAM Toolbox \cite{slamtoolbox2021} for 2D lidar-based mapping and Navigation2 \cite{nav2Macenski} for path planning. The ZedBox handles human detection and tracking using data from the ZED 2 camera, continuously monitoring positions within the camera’s field of view for the predictive model. The MPPI-NIAR algorithm runs on the Intel NUC without GPU acceleration, maintaining a control frequency of up to 25 Hz.

The navigation system operates with two planning layers, where the global and local planners collaborate to navigate the robot around obstacles while adhering to constraints. This process follows the adaptive sub-goal navigation strategy discussed in Section \ref{subsection:subgoal}. A reward map, generated using 2D lidar sensor data, facilitates this functionality. An enhancement involves customizing the obstacle layer within the reward map to use tracking data from the 3D camera, distinguishing between humans and other obstacles. This modification ensures the MPPI-NIAR algorithm treats humans exclusively as dynamic obstacles when detected by the 3D camera, minimizing the risk of misclassification as both dynamic and static entities.

For enhanced safety, the customized reward map includes a fallback mechanism activated during close human-robot interactions, specifically when a human approaches within a 0.2m radius of the robot. In such scenarios, human data is retained in the lidar-based reward map, serving as a redundant safety layer that complements the existing safety protocol relying on 3D stereo camera data. Thus, in these close proximity situations, humans are treated as static obstacles.

\begin{figure*}[!htbp]
    \centering
    \begin{subfigure}[b]{0.4\linewidth}
      \includegraphics[width=\linewidth]{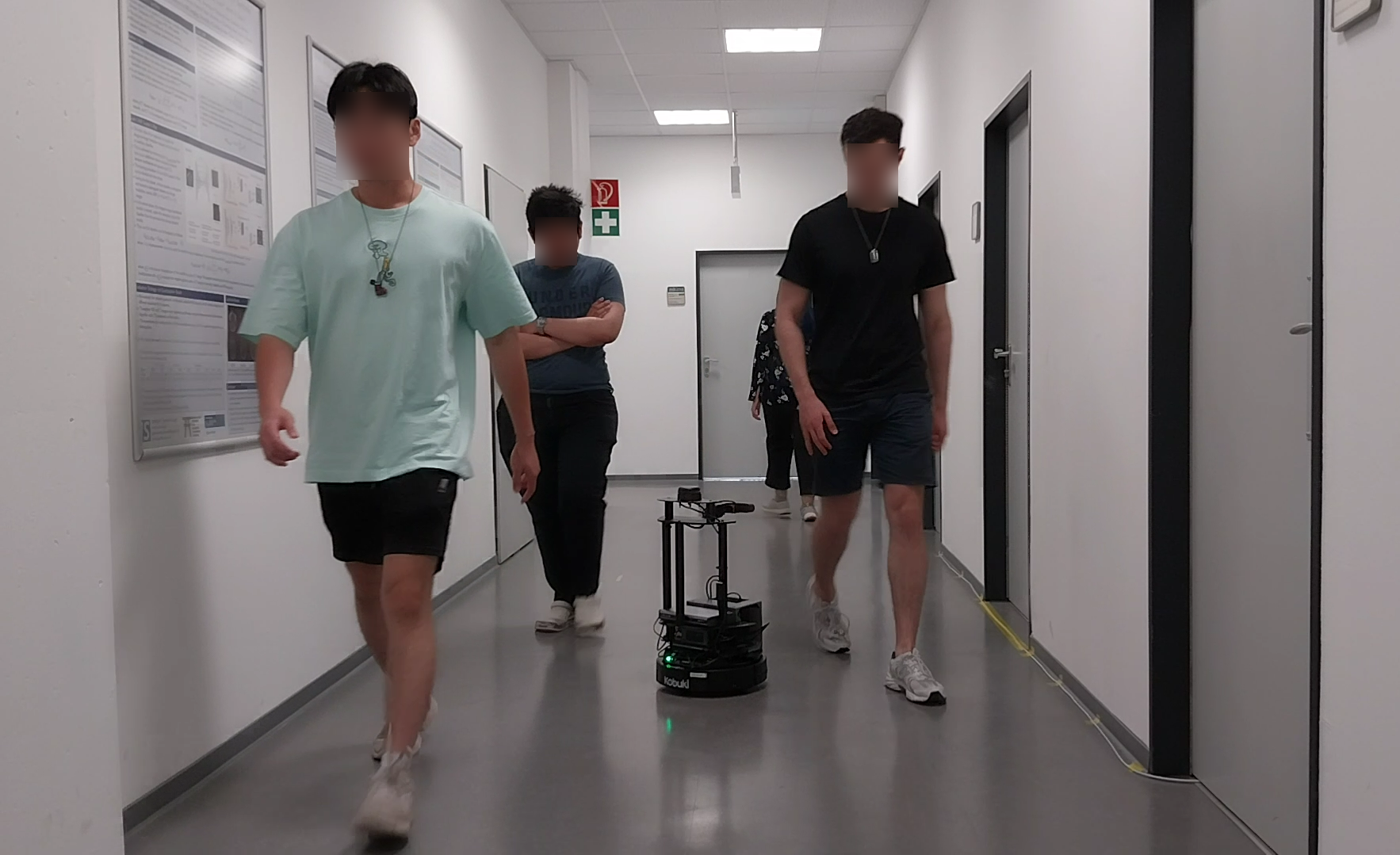}
      \caption{Robot navigating through a crowd.}
      \label{fig:hum2obs}
    \end{subfigure}
    \hfill
    \begin{subfigure}[b]{0.4\linewidth}
      \includegraphics[width=\linewidth]{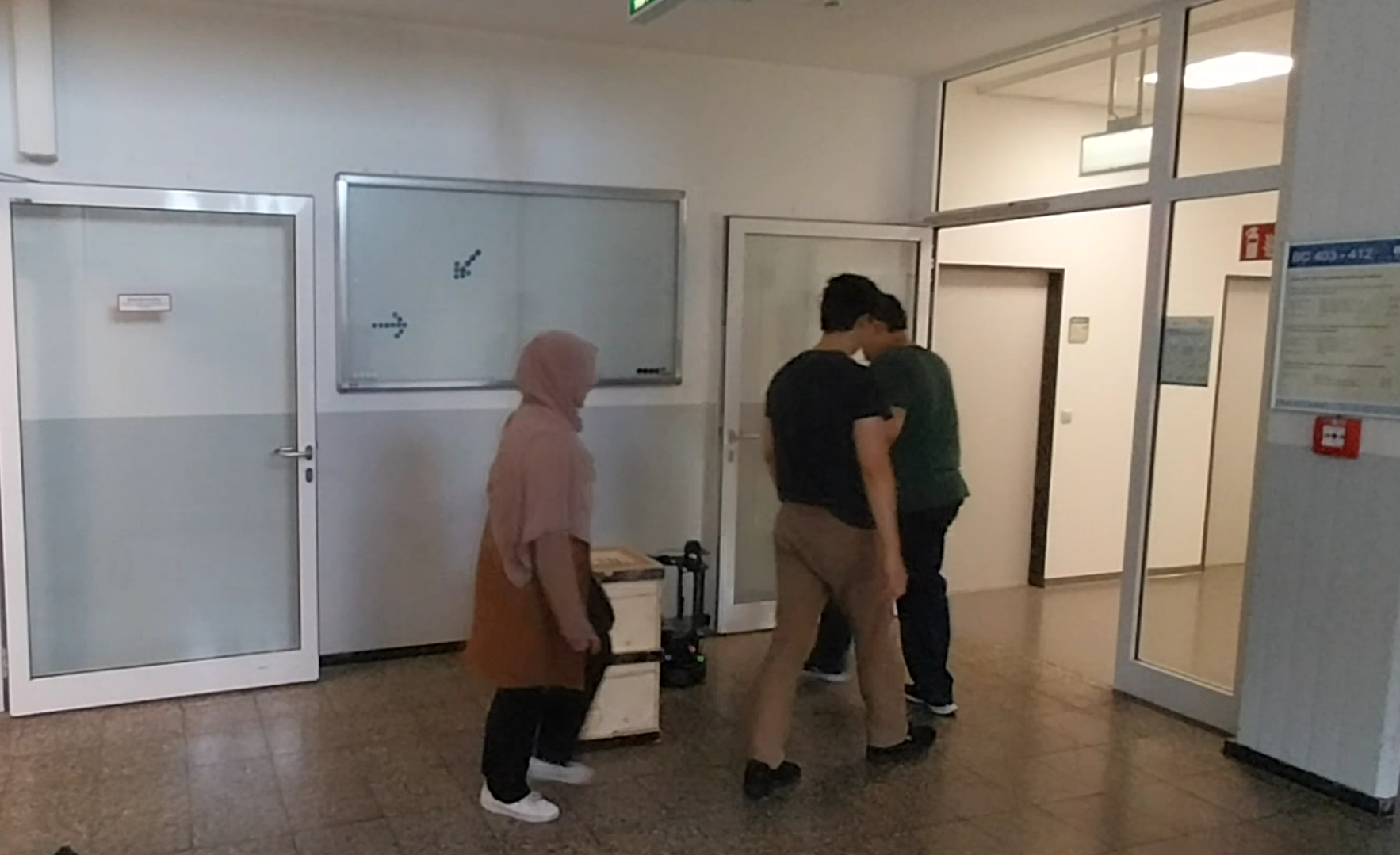}
      \caption{Humans and a static obstacle.}
      \label{fig:hummidobs}
    \end{subfigure}
    \begin{subfigure}[b]{0.4\linewidth}
      \includegraphics[width=\linewidth]{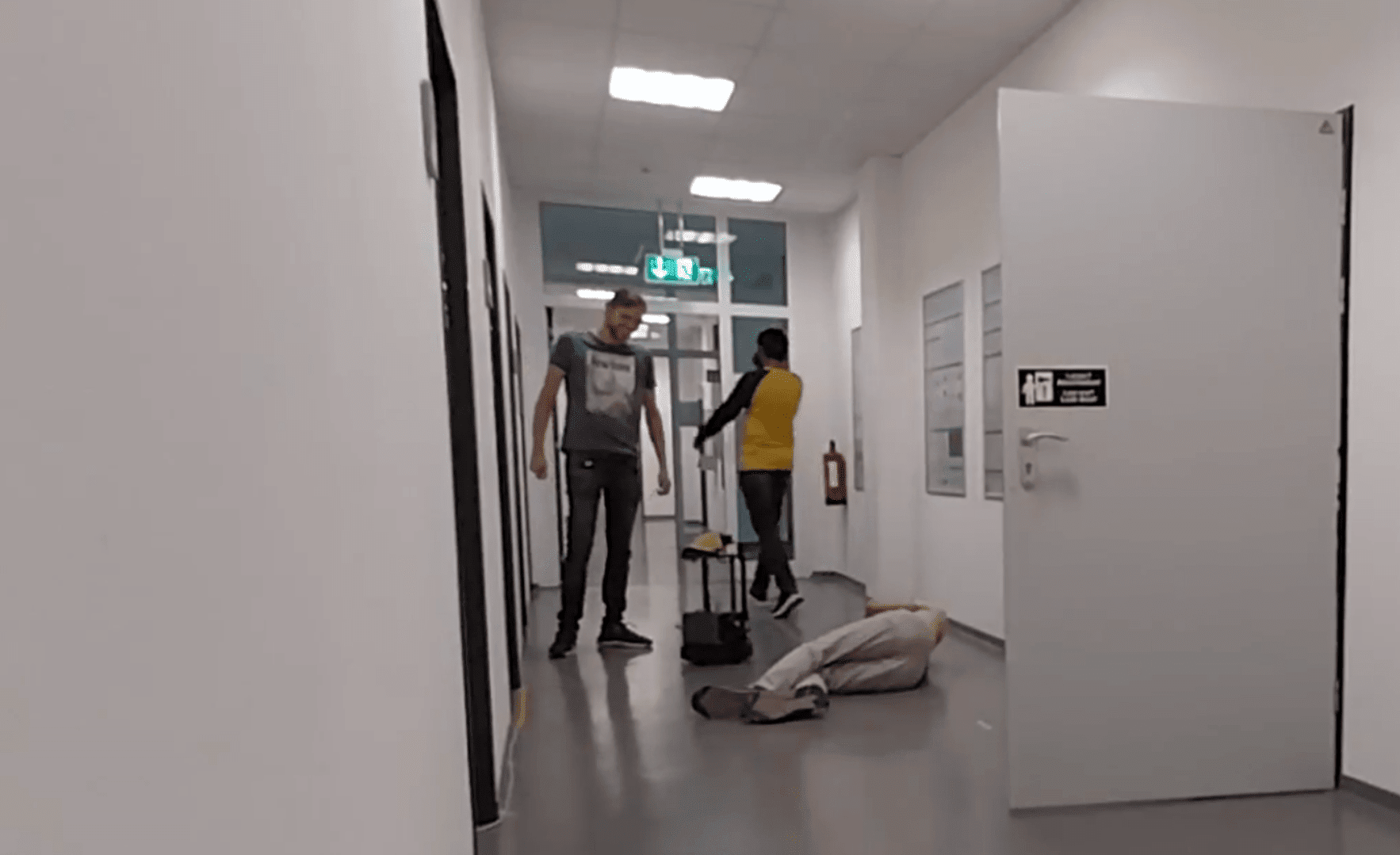}
      \caption{Non-cooperative behavior demonstration.}
      \label{fig:humstatobs}
    \end{subfigure}
    \hfill
    \begin{subfigure}[b]{0.4\linewidth}
      \includegraphics[width=\linewidth]{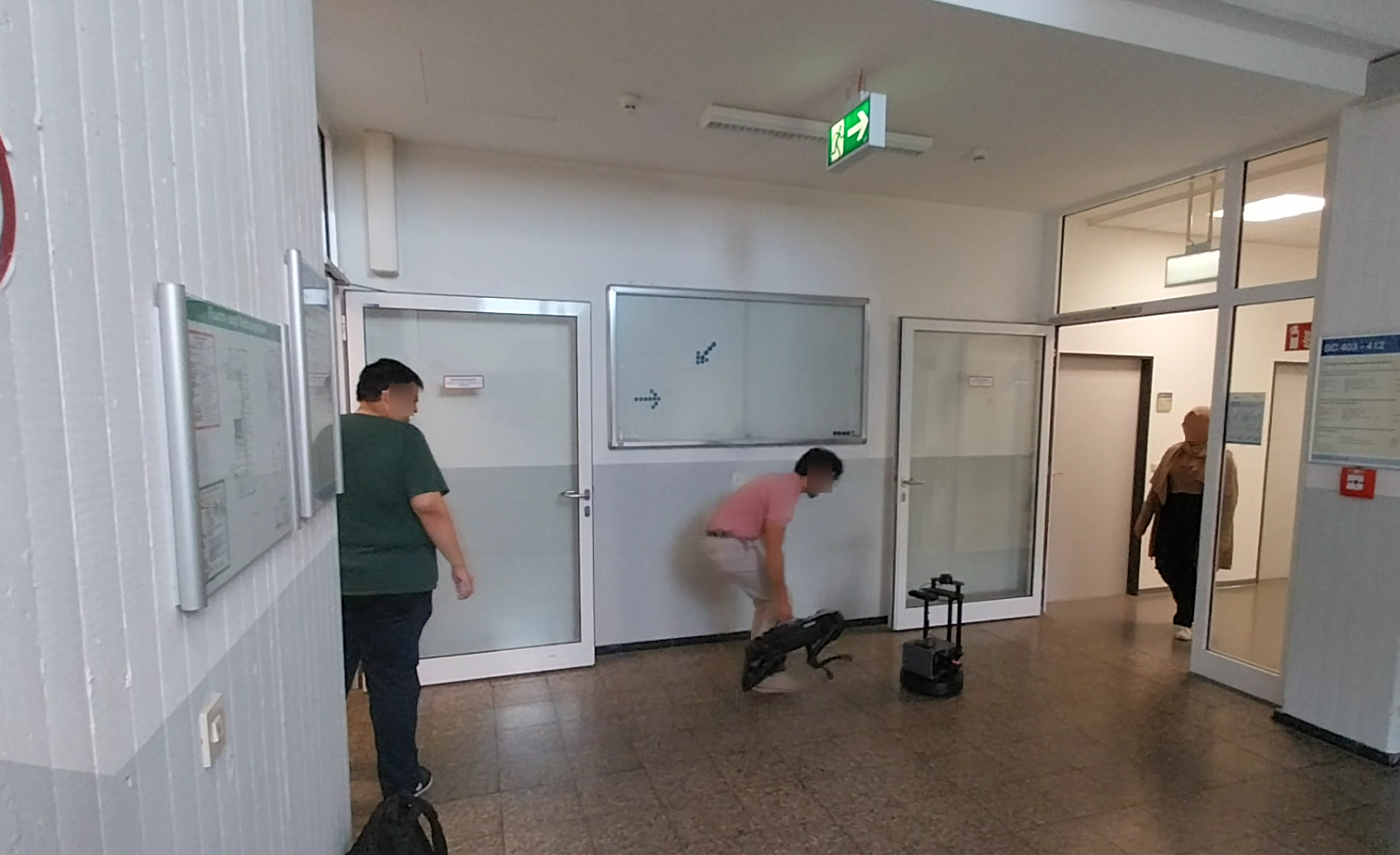}
      \caption{Humans and a dynamic obstacle.}
      \label{fig:humdynobs}
    \end{subfigure}
    \caption{Robot navigation demonstration in various real-world scenarios including humans and obstacles.}
    \label{fig:realevalscene}
  \end{figure*}

The evaluation scenarios for the robotic navigation experiments are illustrated in Figure \ref{fig:realevalscene}. These scenarios emulate various real-world conditions and human interactions that a robot might encounter. The demonstration includes multiple trials at various target locations at the Chair of Intelligent Systems, University of Duisburg-Essen, with interactions involving both cooperative and non-cooperative humans. Volunteers provide qualitative feedback at the end to capture the full spectrum of human-robot interactions.

In the context of the experiments, “cooperative” refers to humans who are aware of the robot’s presence and consciously facilitate its navigation, similar to their interaction with another human. Conversely, “non-cooperative” describes humans who do not make special accommodations for the robot, walking freely with unpredictable movements or even intentional obstructions, testing the system’s resilience and adaptability. Many tests blend these two types of human behavior. Additionally, random static obstacles like chairs and boxes are introduced to add complexity.

Based on the experiments, the robot effectively navigates through corridors with moderate human activity and adeptly maneuvers around both static obstacles and moving humans in confined spaces. The hybrid navigation system, designed with social awareness, ensures the robot remains agile and adaptive. It smoothly navigates complex environments while adhering to the broader goal set by the global plan. This demonstration highlights the planning algorithm’s ability to dynamically model and account for human behavior in real-time scenarios, as shown in Figure \ref{fig:realevalscene}. Recorded test episodes showcasing this functionality are available in the accompanying \href{https://www.youtube.com/watch?v=YsmYvaeuoUg&list=PLILcqCo-rP5eUHchAO_nyRNl0kNgvW5iD&index=1}{videos.}

Although the robot performs admirably in many aspects, feedback from our volunteers suggests several areas for improvement. These include taking calculated risks in obstacle avoidance and enhancing the robot’s responsiveness in crowded situations. Volunteers also suggest improving the robot’s navigation speed, especially in narrow passages and potentially dangerous situations. They often note that the robot sometimes stops when a human is too close. However, this is a safety feature designed to ensure the robot doesn’t navigate too aggressively and to prevent collisions with humans. We propose incorporating more human-like signaling mechanisms, refining algorithms to better deduce intentions, and testing on a more dynamic mobile robot platform. These enhancements could make human-robot interaction safer, more intuitive, and more efficient.

\section{Conclusion}

This study tackles CoBot navigation in crowded environments using goal-conditioned generative models from human crowd videos. These models predict human reactions and select plans that mimic human navigation, and provides a promising direction to enhancing the robot’s goal achievement and human acceptance.

Refining the planning process with goal-conditioning and SIR ensures efficient navigation while respecting social and personal space. SMPC leverages the generative model to produce multiple path samples, managing kinematic and dynamic constraints across different robot platforms.

Experiments with real-world data demonstrate the method’s superiority in safety and efficiency over traditional approaches. Standalone goal-conditioned behavior cloning and offline reinforcement learning struggle due to a lack of interactive data, while online reinforcement learning shows modest but environment-sensitive improvements. Integrating goal-conditioned behavior cloning with SMPC achieves high success rates, robustness, efficiency, and adaptability. Although this work focuses on learning-based and sampling-based planning algorithms, many other approaches should be compared in future studies.

A real-world test with LoCoBot demonstrates compatibility with existing navigation systems, real-time capability, and safety. Future research should address the need for more comprehensive training data to better capture human dynamics and environmental factors, reducing dependency on reward function design and enabling better scalability. Testing the algorithm on more dynamic robotic platforms is also suggested.

\section*{Declarations}

\subsubsection*{Conflict of interest}

The authors declare that they have no conflict of interest.

\subsubsection*{Informed consent} Informed consent was obtained from all individual participants included in the study also for including their data in this paper.

\subsubsection*{Acknowledgments}

We thank Juan Guzman for preparing the RL results, as well as Fatih Özgan and the students who assisted with the real-world demonstration.

\bibliographystyle{sn-basic}
\bibliography{sn-bibliography}

\end{document}